\documentclass[a4paper,fleqn]{cas-dc}

\usepackage[numbers]{natbib}
\usepackage{booktabs}
\usepackage{mathtools}
\usepackage{multirow}
\usepackage{soul}
\usepackage{xcolor}
\usepackage{changepage}
\usepackage{threeparttable}
\usepackage{caption}
\usepackage{subcaption}
\usepackage{pifont}
\usepackage{float}
\usepackage{microtype}

\usepackage[figuresright]{rotating}

\usepackage{algorithm}
\usepackage{amssymb,mathtools,multirow}
\usepackage{amsmath,amssymb,amsfonts}
\newtheorem{assumption}{Assumption}
\newtheorem{theorem}{Theorem}
\newtheorem{lemma}{Lemma}
\usepackage{algpseudocode}
\usepackage{graphicx}
\usepackage{nomencl}
\makenomenclature

\renewcommand{\nompreamble}{The next list describes several symbols that will be later used within the body of the document}

\usepackage{pifont}
\newcommand{\cmark}{\ding{51}}%
\newcommand{\xmark}{\ding{55}}%

\def\tsc#1{\csdef{#1}{\textsc{\lowercase{#1}}\xspace}}
\tsc{WGM}
\tsc{QE}
\tsc{EP}
\tsc{PMS}
\tsc{BEC}
\tsc{DE}

\begin{document}
\let\WriteBookmarks\relax
\def\floatpagepagefraction{1}
\def\textpagefraction{.001}

\shorttitle{Time and Frequency Synergy for Source-Free Time-Series Domain Adaptations}

\shortauthors{M. T. Furqon et~al.}

\title [mode = title]{Time and Frequency Synergy for Source-Free Time-Series Domain Adaptations} 



%
\author[1]{Muhammad Tanzil Furqon}[type=editor]



\ead{muhammad_tanzil.furqon@mymail.unisa.edu.au}



\affiliation[1]{organization={STEM, University of South Australia},
    addressline={Mawson Lakes Boulevard}, 
    city={Adelaide},
    postcode={5095}, 
    country={Australia}}

\author[1]{Mahardhika Pratama}[type=editor]
\ead{dhika.pratama@unisa.edu.au}
\cormark[1]
\cortext[cor1]{Corresponding author}
\author[2]{Ary Shiddiqi}[type=editor]
\ead{ary.shiddiqi@its.ac.id}


\affiliation[2]{organization={Department of Informatics, Institut Teknologi Sepuluh Nopember},
    city={Surabaya},
    postcode={60111}, 
    state={East Java},
    country={Indonesia}}

\author%
[1]
{Lin Liu}
\ead{lin.liu@unisa.edu.au}
\author[1]{Habibullah Habibullah}
\ead{habibullah.habibullah@unisa.edu.au}
\author[1]{Kutluyil Dogancay}
\ead{kutluyil.dogancay@unisa.edu.au}




\begin{abstract}
The issue of source-free time-series domain adaptations still gains scarce research attentions. On the other hand, existing approaches rely solely on time-domain features ignoring frequency components providing complementary information. This paper proposes Time Frequency Domain Adaptation (TFDA), a method to cope with the source-free time-series domain adaptation problems. TFDA is developed with a dual branch network structure fully utilizing both time and frequency features in delivering final predictions. It induces pseudo-labels based on a neighborhood concept where predictions of a sample group are aggregated to generate reliable pseudo labels. The concept of contrastive learning is carried out in both time and frequency domains with pseudo label information and a negative pair exclusion strategy to make valid neighborhood assumptions. In addition, the time-frequency consistency technique is proposed using the self-distillation strategy while the uncertainty reduction strategy is implemented to alleviate uncertainties due to the domain shift problem. Last but not least, the curriculum learning strategy is integrated to combat noisy pseudo labels. Our experiments demonstrate the advantage of our approach over prior arts with noticeable margins in benchmark problems.


\end{abstract}


\begin{highlights}
\item This paper proposes Time Frequency Domain Adaptation (TFDA) to cope with the source free time-series domain adaptation problem.
\item TFDA puts forward the concept of a dual-branch network structure consisting of the time encoder and the frequency encoder. 
\item TFDA proposes the idea of time and frequency consistencies for the source free time-series domain adaptation problem. 
\item Rigorous experiments have been performed where TFDA demonstrate the most encouraging performance over the prior arts.
\end{highlights}


\maketitle

\section{Introduction}

The advent of deep learning \cite{LeCun2015DeepL} has triggered significant research progresses in many applications. The success of deep learning is largely attributed to the i.i.d conditions where the training and deployment phases follow the same distributions but it performs poorly in the case of domain shifts. This problem has motivated the rise of unsupervised domain adaptation (UDA) field \cite{Ganin2014UnsupervisedDA,Kang2019ContrastiveAN} where the objective is to develop a model to generalize well in the unlabelled target domain given labelled samples of the source domain. The source domain and the target domain follow different distributions or are drawn from different domains. Nevertheless, most UDA works rely on the access of labelled samples of the source domain which might be inaccessible because of privacy constraints or limited computational resources. This issue has motivated researchers to explore the source-free domain adaptation (SFDA) topic \cite{Liang2020DoWR,Karim2023CSFDAAC,Litrico2023GuidingPW} which addresses no access of source-domain samples and only exploits a pretrained source-domain model alongside with unlabelled target-domain samples.

\begin{figure}[h]
    \centering
    \includegraphics[width=0.9\linewidth]{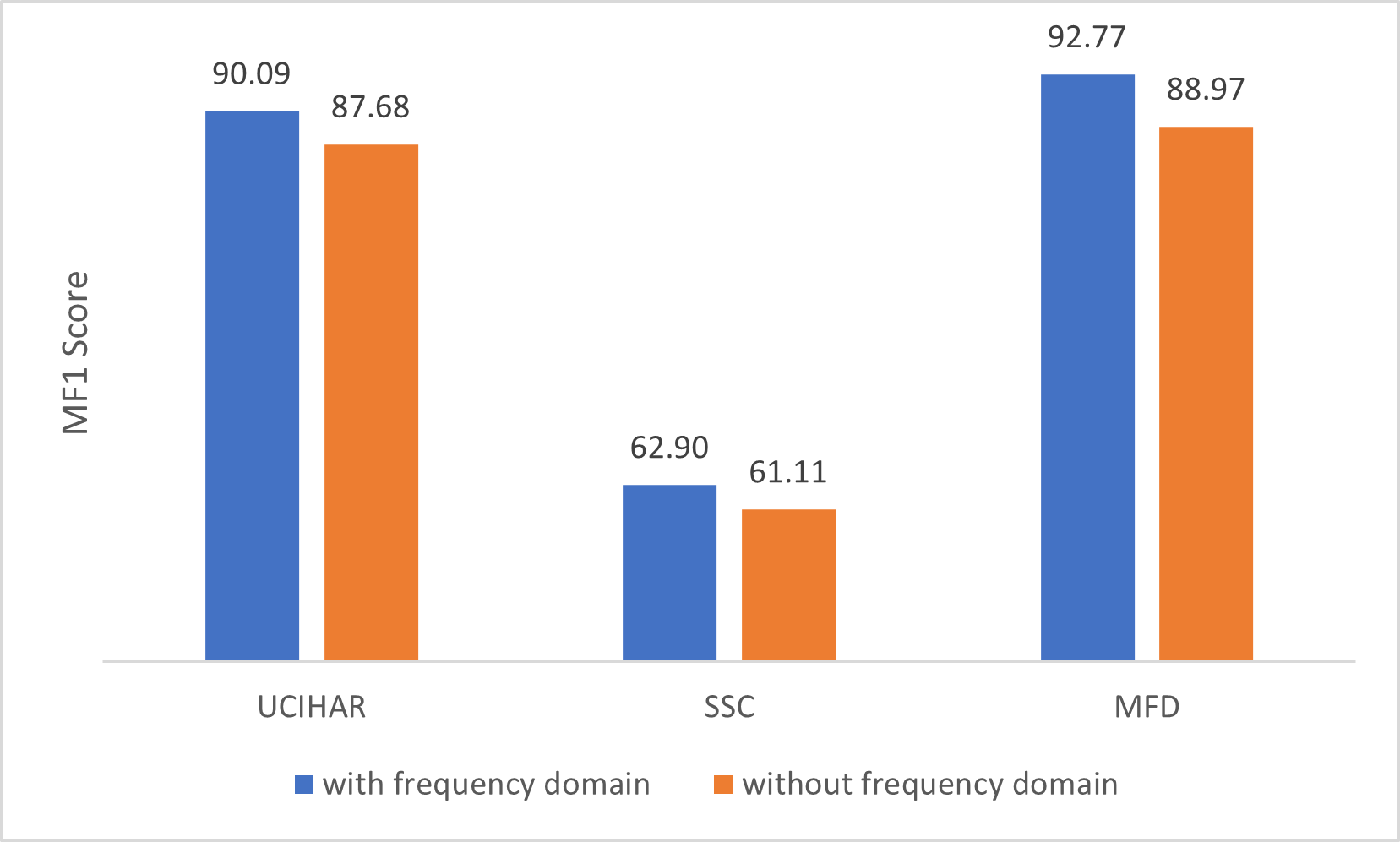}
    \caption{Performance of the model with and without frequency domain.}
    \label{fig:freqNofreq}
\end{figure}

There exist increasing research attentions to study the SFDA topic. In \cite{Liang2020DoWR}, the self-training mechanism is proposed for SFDA using the cluster structure. \cite{Li2020ModelAU} utilizes the generative model to perform SFDA. \cite{Wang2021OntargetA} combines the self-supervised learning approach and the pseudo-labelling approach. Similar approach is put forward in \cite{Chen2022ContrastiveTA} to refine model's predictions with the self-supervised learning strategy. \cite{Karim2023CSFDAAC} designs the curriculum learning approach to prevent early memorization of noisy pseudo-labels while \cite{Litrico2023GuidingPW} presents the loss re-weighting approach based on uncertainties of predictions. All these works focus on image classification problems, possessing no spatio-temporal characteristics compared to that of the time-series classification problems. To the best of our knowledge, the problem of SFDA on time-series data is relatively an uncharted territory where only two works \cite{Zhao2023SourceFreeDA,Ragab2023SourceFreeDA} have been proposed thus far. \cite{Zhao2023SourceFreeDA} addresses the problem but is limited to the case of seizure predictions. In \cite{Ragab2023SourceFreeDA}, the imputation strategy is offered for SFDA on time-series data. These works rely solely on time-domain features and do not yet consider frequency components potentially aiding performance's improvements on time-series domain adaptations \cite{He2023DomainAF}. Figure \ref{fig:freqNofreq} displays the advantage of frequency domain in source-free time-series domain adaptation problem. It is perceived that the use of frequency domain is capable of boosting the numerical results by up to 4\% compared to that of without the frequency domain.  

 Time Frequency Domain Adaptation (TFDA) method is proposed in this paper to handle the absence of source domain data points in UDA given time-series samples. Given a pre-trained source model, the underlying objective is to generalize well over unlabelled target-domain samples. TFDA is built upon a dual branch network structure taking into account both time and frequency features. That is, it possesses a time encoder and a frequency encoder processing time and frequency features independently. The final output is aggregated based on their confidence degrees. It adopts the self-training mechanism taking into account the neighborhood structure of target domain samples. That is, a pseudo-label is derived from those of neighboring samples given that adjacent samples should share the same class in respect to the smoothness constraint. To this end, the contrastive learning framework in both time and frequency domains is designed to pull the same-class samples together while pushing away different-class samples. In addition, the time-frequency consistency is derived using the self-distillation concept because the time and frequency representations should be close to each other. That is, the domain adaptation step is achieved by aligning the temporal and frequency features. Since the presence of domain shifts between the source and target domains result in uncertainties of predictions, the uncertainty reduction approach is put forward. Last but not least, the curriculum learning strategy is proposed to overcome the noisy pseudo label problem. Such learning mechanism prevents learning of unreliable samples in the early stage of the training process causing early memorization of noisy pseudo labels. Easy-to-learn samples indicating clean labels are learned first before progressively learning hard samples. 

This paper proposes the following major contributions:
\begin{itemize}
    \item TFDA proposes the concept of a dual branch network structure fully utilizing both time and frequency components. That is, the time encoder and the frequency encoder are decoupled with their own classifiers. The final predictions are aggregated from the confidence concept. This strategy allows rich information to be considered in deriving final predictions.
    \item The contrastive learning strategy is developed using the pseudo label information to avoid features of the same class to be pushed apart and is carried out in both time and frequency domains to make valid neighborhood assumptions. The negative pair exclusion strategy is carried out to prevent noisy pseudo labels taking into account the history of pseudo labels.
    \item The time-frequency consistency is proposed using the concept of self-distillation. That is, time representations and frequency representations must be close to each other \cite{Zhang2022SelfSupervisedCP}. It differs from \cite{Zhang2022SelfSupervisedCP} where the concept of self-distillation is applied rather than the triplet loss. 
\end{itemize}
Rigorous numerical and theoretical studies have been carried out where TFDA is compared with various UDA and SFDA approaches. TFDA demonstrates the most encouraging performances outperforming prior arts with significant margins. Source codes of TFDA is shared publicly in \url{https://github.com/furqon3009/TFDA} to ensure reproducibility and convenient further study.





\section{Related Works}

\subsection{Time Series Domain Adaptation}
The problem of domain shifts on time-series data has been studied where it aims to generalize well over unlabelled samples of the target domain given labelled samples of the source domain. The underlying challenge lies in the temporal nature of time-series samples besides the distribution shifts of the source domain and the target domain. Existing works can be categorized into two groups: adversarial-based approach and discrepancy-based approach. AdvSKM \cite{Liu2021AdversarialSK} presents a discrepancy-based approach integrating spectral kernel to address the temporal dependencies on top of the MMD approach. SASA \cite{Cai2020TimeSD} utilizes the association structure across the two domains for time-series UDA. The adversarial-based method plays the adversarial game to minimize the domain gap. Such approach is exemplified by CoDATS \cite{Wilson2020MultiSourceDD} incorporating the adversarial learning for multi-source human-activity recognition tasks. DAATTN \cite{Jin2021DomainAF} combines the adversarial training strategy and the attention sharing mechanism. SLARDA \cite{Ragab2021SelfsupervisedAD} presents an autoregressive approach for adversarial training. In \cite{Furqon2024MixupDA}, the concept of mixup domain adaptation is developed to handle domain adaptations of aircraft engines.  

These works depend on the access of source-domain data, often unavailable in several practical applications due to the privacy constraints or the storage limitations. TFDA works with only a pretrained source model with the absence of any source-domain data. The pretrained source model is adapted using only unlabelled target-domain data.

\subsection{Source-Free Domain Adaptation}
Source-free domain adaptation (SFDA) aims to address the issue of privacy in unsupervised domain adaptation (UDA) where source-domain data are unavailable when performing domain adaptations. \cite{Liang2020DoWR} puts forward the self-training mechanism using the cluster's structure. \cite{Li2020ModelAU} presents the generative model to perform SFDA. The self-supervised learning strategy is offered in \cite{Chen2022ContrastiveTA}. \cite{Karim2023CSFDAAC} devises the curriculum learning strategy to prevent early memorization of noisy pseudo-labels while \cite{Litrico2023GuidingPW} is based on the loss re-weighting strategy. These works are designed for visual applications which do not possess any temporal properties. 

To date, only two methods have been proposed in the literature to deal with source-free time-series domain adaptation. \cite{Zhao2023SourceFreeDA} utilizes the Gaussian Mixture Model (GMM) for seizure data. In \cite{Ragab2023SourceFreeDA}, the time-series imputation strategy is proposed. These works are solely based on temporal features without taking advantages of frequency features which can improve model's generalization on time-series data \cite{He2023DomainAF}.


\section{Problem Formulation}

Given a pretrained source model $g_{\phi}(f_{\theta}(x))$ where $f_{\theta}(.):\mathcal{X}\rightarrow\mathcal{Z}$ is a feature extractor mapping the input space to the latent space while $g_{\phi}(.):\mathcal{Z}\rightarrow\mathcal{Y}$ is a classifier converting the feature space into the output space, the objective of SFDA is to generalize well over the unlabelled target domain $\mathcal{T}=\{(x_i^{T})\}_{i=1}^{N_T}$ where $N_T$ denotes the number of unlabelled samples in the target domain, $x^{T}\in\mathcal{X}^{T},y^{T}\in\mathcal{Y}^{T},\mathcal{X}^{T}\times\mathcal{Y}^{T}\in\mathcal{D}^{T}$. The source model $g_{\phi}(f_{\theta}(.))$ is pretrained using the labelled samples of the source domain $\mathcal{S}=\{(x_i,y_i)\}_{i=1}^{N_S}$ where $N_S$ stands for the number of labelled source-domain samples, $x^{S}\in\mathcal{X}^{S},y^{S}\in\mathcal{Y}^{S},\mathcal{X}^{S}\times\mathcal{Y}^{S}\in\mathcal{D}^{S}$. The source domain and the target domain possess the domain shift problem $\mathcal{D}^{S}\neq\mathcal{D}^{T}$ due to different marginal distributions $P^{S}(x)\neq P^{T}(x)$ but share the same label space $\mathcal{Y}^{S}=\mathcal{Y}^{T}$. We limit our scope of study to a closed-set problem. $\mathcal{D}^S,\mathcal{D}^T$ stand for the source domain and the target domain respectively while $P^{S}(x),P^{T}(x)$ denote the marginal distribution of the source domain and the target domain respectively. Because of the issue of privacy, the domain adaptation phase is done with the absence of any source-domain data $\mathcal{S}$. That is, $\mathcal{S}$ is discarded once used in the pre-training phase.

\section{Method}
\begin{figure}[h]
    \centering
    \includegraphics[width=0.9\linewidth]{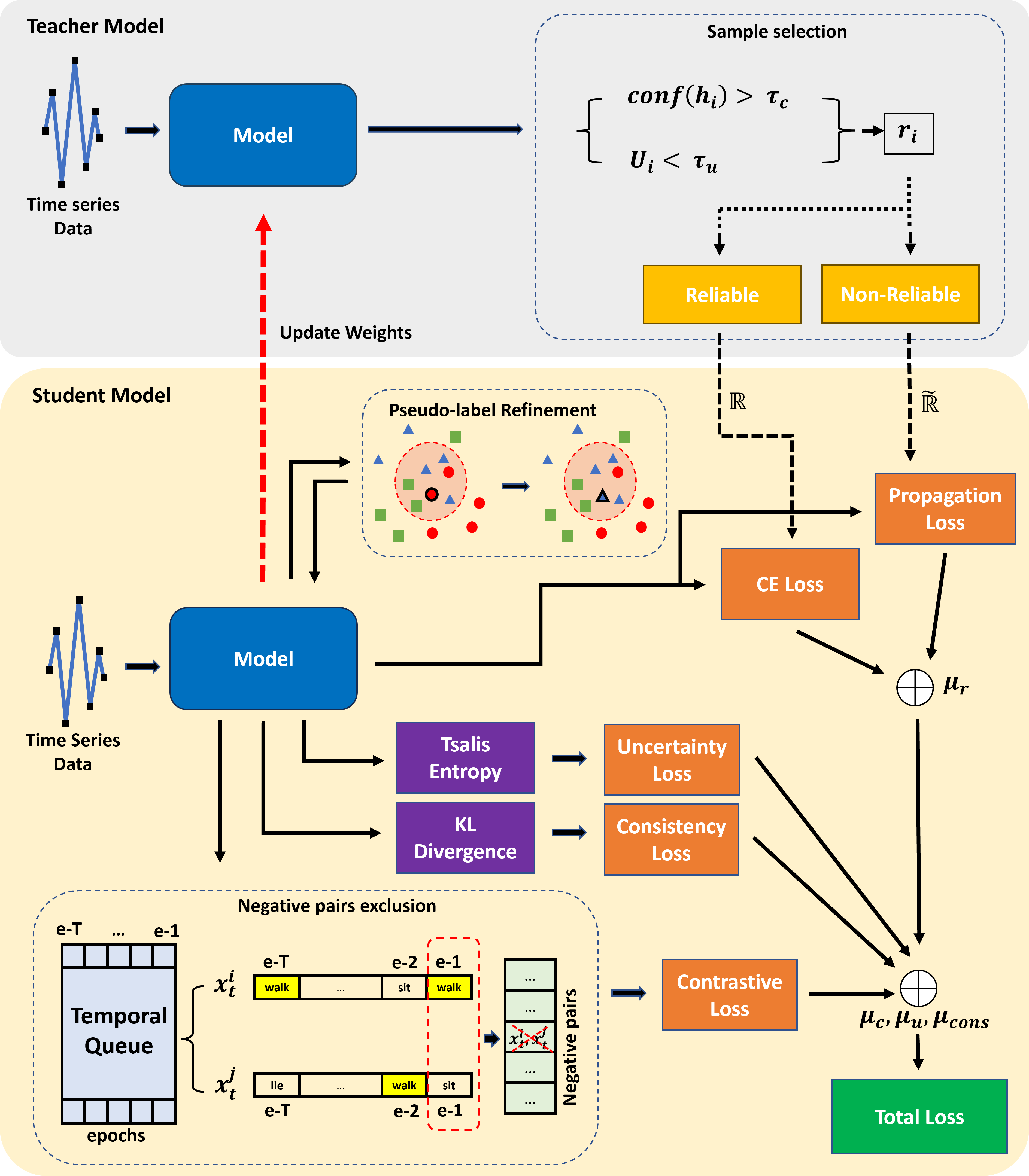}
    \caption{TFDA algorithm starts with the neighborhood pseudo-labelling strategy. After that, data samples are divided into two parts: reliable and non-reliable parts learned differently with class-balanced cross-entropy loss and label propagation loss under the curriculum learning framework. In addition, the contrastive learning framework, the self-distillation learning framework and the uncertainty learning framework are integrated.}
    \label{fig:sftsda}
\end{figure}

TFDA is developed from the teacher-student architecture as shown in Fig. \ref{fig:sftsda} under a dual-branch network structure processing temporal and frequency features independently. That is, the exponential moving average strategy is applied to tune the teacher's model based on the parameters of the student's model. First, the student model is trained to minimize the teacher model losses. This strategy is implemented because a weighted-average model normally performs better than the final model. It starts with the neighborhood pseudo-labelling strategy followed by the sample selection strategy to determine reliable and unreliable samples learned differently under the roof of the curriculum learning. The contrastive learning strategy is integrated to satisfy the smoothness constraint while the uncertainty learning strategy is incorporated to overcome the domain shift problem. The contrastive learning approach takes place in both time domain and frequency domain. The domain adaptation mechanism is attained by aligning the time components and the frequency components via the self-distillation technique. 

\subsection{Dual Branch Structure}
\begin{figure}[h]
    \centering
    \includegraphics[width=0.9\linewidth]{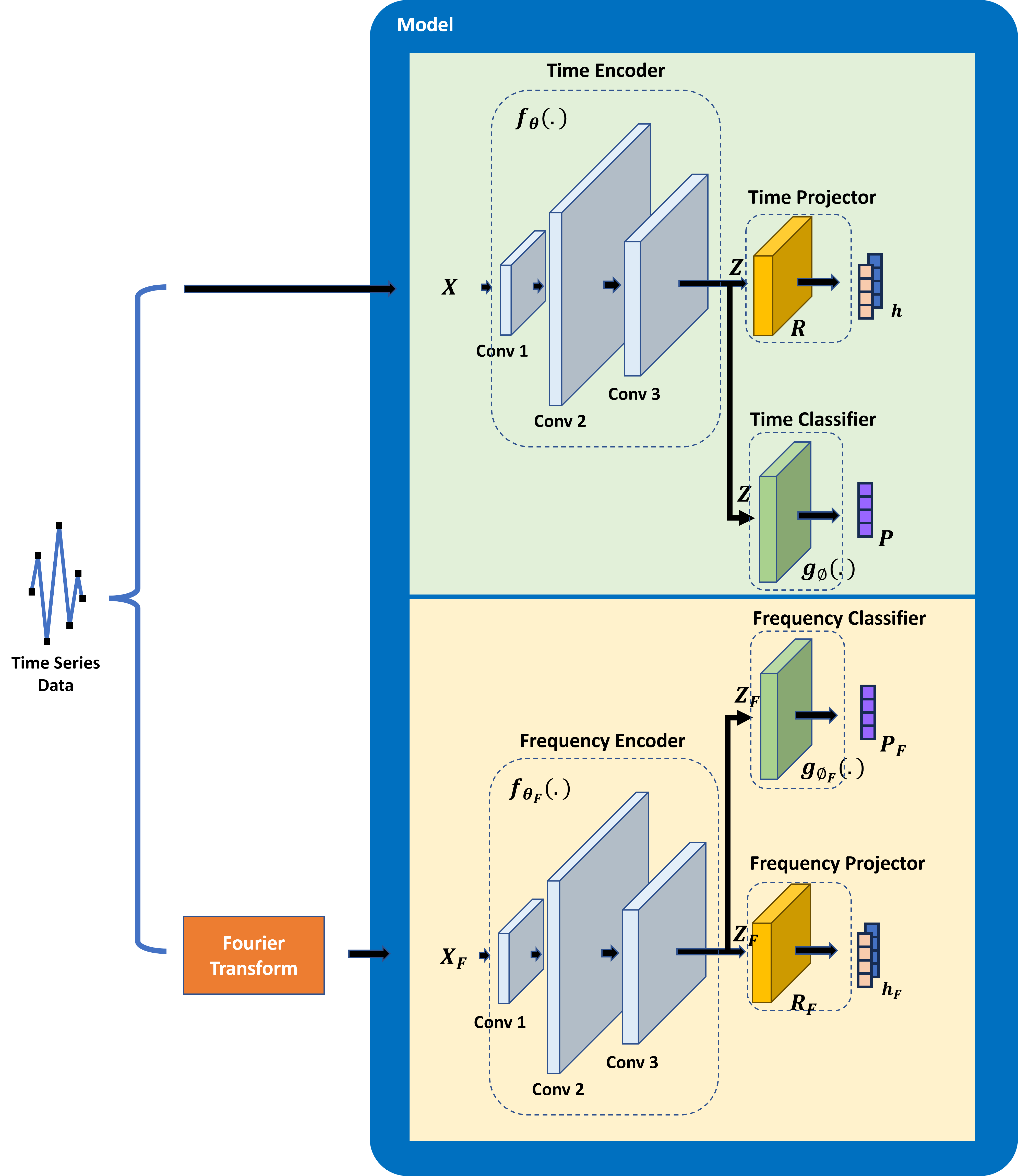}
    \caption{TFDA architecture: it comprises the temporal encoder processing the temporal features and the frequency encoder mining the frequency features. The Fourier transform is applied to extract the frequency spectrum. Each encoder is assigned with its own classifier where the final output is aggregated based on the confidence of their predictions.}
    \label{fig:sftsdaArch}
\end{figure}

TFDA is developed from a dual branch network structure where each branch mines the time features or the frequency features independently. The motivation of the use of frequency features lies in the fact that the frequency features are representations of the same signal as the features, thereby offering complementary information. That is, it possesses the time encoder $f_{\theta}(.):\mathcal{X}\rightarrow\mathcal{Z}$ and the frequency encoder $f_{\theta_f}(.):\mathcal{X}_{F}\rightarrow\mathcal{Z}_{F}$ where $x_{F}$ is a frequency input $x$ using the Fourier transform for transformation. Each encoder is coupled with its own classifier $g_{\phi}(.):\mathcal{Z}\rightarrow\mathcal{P}$ and $g_{\phi_{f}}(.):\mathcal{Z}_{F}\rightarrow\mathcal{P}_{F}$. The final prediction is drawn from the weighted mixture of the two network outputs \cite{Wang2023FewShotCS}. 
\begin{equation}
    p_{i}=\alpha p(y_i=c|x_i) + \beta p_{F}(y_i=c|x_{i}^{f})
\end{equation}
where $\alpha=\frac{\arg\max_{c\in[0,C]}\alpha_c}{\arg\max_{c\in[0,c]}\alpha_c+\arg\max_{c\in[0,c]}\beta_c}$ and \newline $\beta=\frac{\arg\max_{c\in[0,C]}\beta_c}{\arg\max_{c\in[0,c]}\alpha_c+\arg\max_{c\in[0,c]}\beta_c}$. $\alpha,\beta$ respectively stand for the soft labels produced by $p,p_{F}$. This strategy allows rich information of the time and frequency spectrum to be considered when inducing the final predictions. Fig. \ref{fig:sftsdaArch} portrays the network architecture which features a dual-branch network structure. 

\subsection{Pseudo-labelling Strategy}
Inspired by \cite{Litrico2023GuidingPW}, TFDA applies the concept of neighborhood aggregation when deriving pseudo labels. That is, similar samples should share the same label and be close in the feature space whereas dissimilar samples should possess distinct label and be far away in the latent space. It is attained via the contrastive learning strategy pulling similar samples while pushing dissimilar samples. The use of sample groups in deriving pseudo labels reduces the risk of noisy pseudo label better than based on a single sample. 

For a target sample $x_t$ and a weak augmentation $t_{wa}$ of the distribution $\mathcal{T}_{wa}$, a feature vector $z=f_{\theta}(t_{wa}(x_t))$ is elicited and used to find neighboring samples. The weak augmentation applies the jitter-and-scale strategy \cite{Eldele2021TimeSeriesRL} where random variations are added to the signals and its magnitudes are scaled up. The pseudo label of $x_t$ is obtained from aggregations of neighbors:
\begin{equation}
    \hat{p}_i^{c}=\frac{1}{K}\sum_{i\in\mathcal{I}}p_{i}^{'(c)}
\end{equation}
where $\mathcal{I}$ denotes the set of indices of the neighbors and $K$ stands for the number of neighbors. The argmax operator is applied to obtain the refined pseudo label. 
\begin{equation}
    \hat{y}_{t}=\arg\max_{c}\hat{p}_{t}^{(c)}
\end{equation}
We apply the memory bank $B$ of length $M$ which stores the candidates of neighbors $\{z_j^{'},p_{j}^{'}\}_{j=1}^{M}$. The neighbors are selected by the cosine distance between the target feature $x_t$ and those of the memory bank where the $K$ features with the smallest distances are selected as neighbors.
\subsection{Sample Selection}
TFDA applies the sample selection strategy to prevent the influence of noisy pseudo-labels. That is, reliable labels are learned confidently since the beginning of the training process whereas unreliable labels are initially avoided. Motivated by \cite{Karim2023CSFDAAC}, we put forward the prediction confidence and uncertainty as a reliable measure of pseudo label accuracy. Nevertheless, the problem of domain shifts bias the model's predictions and no source data samples are available to estimate the domain's discrepancy. To this end, an augmentation policy is designed to indicate the virtual distribution shift across target domain data where prediction variance or uncertainty over augmented distributions offers the close estimate of the domain shift.

Suppose that $h_i=g_{\phi}(f_{\theta}(x_i))$ and $conf(h_i)=\max_{c}h_i$ where $C$ stands for the number of target classes, The reliability score is derived as follows:
\begin{equation}
    r^{i}=  \begin{cases} 1, \quad conf(h_i)\geq\tau_{c}\quad \&\quad u_{i}\leq\tau_u\\
    0, \quad otherwise
    \end{cases}
\end{equation}
where $u_i$ denotes the uncertainty of the $i-th$ sample calculated as the standard deviation of the prediction confidence over $L$ augmentations  $u_i=std\{conf(h_i^{l})\}_{l=1}^{L}$ and $L$ stands for the number of augmented samples. $\tau_c,\tau_u$ stand for the confidence and uncertainty thresholds simply set as the average over a batch $\tau_c=\frac{1}{B}\sum_{i=1}^{B}conf(h_i)$ and $\tau_u=\frac{1}{B}\sum_{i=1}^{B}u_i$ where $B$ denotes the mini-batch size.

\subsection{Self-Distillation Strategy}
The reliability score $r^{i}$ allows segregation of data samples into two groups: reliable group $\mathbb{R}=\{(x_i,y_i):r_i=1\}_{i=1}^{B}$ and non-reliable group $\mathbb{\tilde{R}}=\{(x_i,y_i):r_i=0\}_{i=1}^{B}$ to be learned with different paces. We apply the top-2 confidence score from $\mathbb{\tilde{R}}$ to guarantee diversity of the reliable group $\mathbb{R}$. The class-balanced cross-entropy loss function $\mathcal{L}_{ce}$ is implemented to learn the reliable group $\mathbb{R}$ whereas the non-reliable group $\mathbb{\tilde{R}}$ is learned by the label propagation loss $\mathcal{L}_{lp}$:
\begin{equation}\label{eq:self-learning}
    \mathcal{L}_{lp}=\frac{1}{2|\mathbb{\tilde{R}}|}\sum_{i=1}^{|\mathbb{\tilde{R}}|}||g_{\phi}(f_{\theta}(x_i))-\hat{y}_{i}||_2
\end{equation}
where the label information is transferred from the reliable group $\mathbb{R}$ to the non-reliable group $\mathbb{\tilde{R}}$ due to the transductive property $\mathcal{L}_{lp}$.

\subsection{Contrastive Learning}
The contrastive learning strategy \cite{Zhang2022SelfSupervisedCP,Litrico2023GuidingPW} is adopted to pull similar samples, positive pairs, together while pushing away dissimilar samples, negative pairs. This strategy is required because of the neighborhood pseudo-labelling strategy where adjacent samples should share the same label. Two strong augmentations $t_{sa},t_{sa}^{'}\in\mathcal{T}_{sa}$ are selected to generate two different views of a sample $t_{sa}(x),t_{sa}^{'}(x)$ encoded to be query $q=f_{\theta}(t_{sa}(x))$ and key $k=f_{\theta}(t_{sa}^{'}(x))$ through the encoder $f_{\theta}$. The keys and queries construct the positive pair, while the negative pairs are created from a queue $Q$ storing key features of each mini-batch $\{k_i\}_{i=1}^{N}$,i.e., other samples in the mini-batch as per \cite{He2019MomentumCF}. The strong augmentation is resulted from the permutation-and-jitter strategy \cite{Eldele2021TimeSeriesRL}. That is, a signal is split into a random number of segments with a maximum of $M$ and randomly shuffled.  

We consider pseudo-labels when performing the contrastive learning mechanism to avoid features having the same category to be pushed away. This problem is addressed in \cite{Chen2022ContrastiveTA} with the negative pair exclusion strategy where the negative pair is masked out if the two samples possess the same pseudo-label. Nevertheless, this strategy ignores the noisy pseudo label problem undermining the negative pair assumption. We follow \cite{Litrico2023GuidingPW} which takes into account the history of pseudo labels. That is, the queue $Q$ is converted into a temporal queue also memorizing the pseudo labels $\{\hat{y}_{i}\}_{i=1}^{T}$ across the past $T$ epochs. The pair is excluded if it has the same pseudo-label at least once over the past $T$ epochs. The contrastive learning is achieved by minimizing the following InfoNCE loss function. 
\begin{equation}
    \mathcal{L}_{cl}=-\log{\frac{\exp{(q.k_{+}/\tau)}}{\sum_{j\in\mathcal{N}_{q}}\exp{(q.k_j/\tau)}}}
\end{equation}
\begin{equation}
    \mathcal{N}_{q}=\{j|\hat{y}_{j}^{i}\neq\hat{y}^{i},\forall j\in\{1,...,N\},\forall i\in\{1,...,T\}\}
\end{equation}
where $\mathcal{N}_{q}$ is the set of indices in $Q$ that never possess the same pseudo label with the query sample in the past $T$ epochs. $q,k$ respectively stand for the query and the key 

Existing SFDA methods focus solely on the time domain and overlooks the frequency domain \cite{Zhang2022SelfSupervisedCP} potentially boosting the performances given time-series data. As a matter of fact, the time and frequency domains are different views of the same data which should be invariant regardless of the time series distributions. We are motivated to extract a general characteristic preserved across the time-series datasets. 

The frequency spectrum $x_i^{f}$ is produced by applying a transform operator to the time-domain sample $x_i$. The frequency encoder $f_{\theta_{f}}$ is applied to embed the frequency spectrum $x_i^{f}$. The augmentation procedure follows \cite{Zhang2022SelfSupervisedCP} where the frequency spectrum is perturbed by adding or removing the frequency components. Via the two augmentation methods $t_{sa},t_{sa}^{'}\in\mathcal{T}_{sa}$, the query $q^{f}=f_{\theta_{f}}(t_{sa}(x^{f}))$ and the key $k^{f}=f_{\theta_{f}}(t_{sa}^{'}(x^{f}))$ are generated to produce two different views of the frequency spectrum $x^{f}$ and form the positive pair. As with the time-domain features, the negative pair exclusion strategy is applied. The frequency contrastive learning is expressed mathematically as follows:
\begin{equation}
    \mathcal{L}_{cl}^{f}=-\log{\frac{\exp{(q^{f}.k_{+}^{f}/\tau)}}{\sum_{j\in\mathcal{N}_{q}^{f}}\exp{(q^{f}.k_j^{f}/\tau)}}}
\end{equation}
\begin{equation}
    \mathcal{N}_{q}^{f}=\{j|\hat{y}_{j}^{i}\neq\hat{y}^{i},\forall j\in\{1,...,N\},\forall i\in\{1,...,T\}\}
\end{equation}
where $\tau$ is the temperature constant. Given the time embedding via the time encoder $f_{\theta}(.)$ and the frequency embedding via the frequency encoder $f_{\theta_{f}}(.)$, the next task is to perform the contrastive learning between time and frequency features $\mathcal{L}_{cl}^{tf}$. This is done by projecting their representations into a joint time-frequency space through projectors $R(.)$ and $R_{f}(.)$ mapping the frequency or time embedding into the joint embedding. That is, the query and the key are constructed as $q=R(f_{\theta}(x))$ and $k=R_{f}(f_{\theta_{f}}(x_{f}))$. The time-frequency contrastive learning is performed similarly as that the time or frequency contrastive learning considering the presence of pseudo labels with the negative pair exclusion strategy. The main distinction lies in the use of joint embedding $R(.)$ and $R_{f}(.)$ to produce the time-frequency features. 
\begin{equation}
\mathcal{L}_{cl}^{tf}=-\log{\frac{\exp{(q.k_{+}^{f}/\tau)}}{\sum_{j\in\mathcal{N}_{q}}\exp{(q.k_j^{f}/\tau)}}}
\end{equation}
This implies time and frequency features of the same classes are pulled together while pushing apart those of different class. That is, the time and frequency features should be close to each other because they represent the same sample. Because of the absence of ground truth, the supervision signal is generated by the pseudo-labelling process with the negative pair exclusion method. The overall contrastive learning strategy is formulated:
\begin{equation}\label{eq:cl}
    \mathcal{L}_{CL}=\alpha_{1}(\mathcal{L}_{cl}^{f}+\mathcal{L}_{cl})+\alpha_{2}\mathcal{L}_{cl}^{tf}
\end{equation}
where $\alpha_1,\alpha_2$ stand for the tradeoff constants fixed at $0.5$ in our simulations meaning that both terms possess equal impacts. 

\subsection{Time Frequency Consistency}
The consistency between the time component and the frequency component is attained via the self-distillation technique and done in the output space. That is, the outputs of frequency encoder and time encoder are forced to be consistent. The following consistency loss is formulated using the KL divergence. 
\begin{equation}
\begin{split}
    \mathcal{L}_{cons}=\mathcal{D}_{KL}(g_{\phi}(f_{\theta}(x))||g_{\phi_{f}}(f_{\theta_{f}}(x_{f})))+\\\mathcal{D}_{KL}(g_{\phi_{f}}(f_{\theta_{f}}(x_{f}))||g_{\phi}(f_{\theta}(x)))
\end{split}
\end{equation}
where $D_{KL}(a||b)=\sum_{x\in\mathcal{X}}a(x)\log{(\frac{a(x)}{b(x)}})$. It differs from \cite{Zhang2022SelfSupervisedCP} where the consistency is maintained in the embedding space. Our approach constructs different architectures between the frequency component and the time component where the outputs of the two architectures are matched at the end.

\subsection{Uncertainty Reduction Strategy}
In realm of SFDA, the domain shift causes uncertainties of model predictions \cite{Zhao2023SourceFreeDA} and minimization of uncertainties leads to improved robustness. Due to the domain shift, a model tends to learn samples with high confidence but underutilize those having high uncertainties. We apply a modified Tsalis entropy approach \cite{Xia2022PrivacyPreservingDA,Zhao2023SourceFreeDA} to minimize uncertainties. 
\begin{equation}
    \mathcal{L}_{ul}=-\frac{1}{(a-1)}\frac{1}{C}\sum_{i=1}^{n_t}\sum_{c=1}^{C}\frac{\eta_i(h_i^{k})^{a}}{\beta_k}
\end{equation}
\begin{equation}    \beta_k=\sum_{i=1}^{n_t}h_i^{k}
\end{equation}
\begin{equation}
    \eta_i=\frac{n_t[1+\exp{(-E(h_i))}]}{\sum_{j=1}^{n_t}[1+\exp{(-E(h_j))}]}
\end{equation}
where $a$ denotes the exponent parameter of outputs empirically set to $2$ and $h_i=[h_i^{1},...,h_i^{C}]$ is the output probability. $E(h_i)=-\sum_{c}^{C}h_i^{c}\log{h_i^{c}}$ is the entropy of $x_i$. $C$ denotes the number of classes. This loss function modifies the Tsallis entropy in two ways: 1) the influence of samples with high entropy is reduced. Hence, weights are assigned to data points according to their own entropy levels via the Laplace smoothing; 2) to alleviate the potential class imbalance problem in prediction, the output of class $c$ is normalized.

\subsection{Overall Loss Function}
The overall loss function is formalized as follows:
\begin{equation}
\begin{split}
    \mathcal{L}_{all}=\mu_r\mathcal{L}_{ce}+(1-\mu_{r})\mathcal{L}_{lp}+\mu_{c}\mathcal{L}_{CL}+\\\mu_{cons}\mathcal{L}_{cons}+\mu_{u}\mathcal{L}_{ul}
\end{split}
\end{equation}
where $\mu_r,\mu_{c}, \mu_{cons},\mu_{u}$ are trade-off coefficients controlling the pace of learning process. We adopt the curriculum learning principle to steer the tradeoff coefficients. 

Since the performance of a weighted-average model over time steps is generally better than the final model \cite{Raychaudhuri2023PriorguidedSD}, the mean-teacher framework is implemented here. That is, it involves the teacher-student approach initialized using the source model. Suppose that $\Theta=\{\theta,\phi\}$, the parameters of the student model $\Theta^{st}$ are adjusted by back-propagating the losses incurred by the teacher model. The parameters of the teacher model $\Theta^{te}$ is adapted using the exponential moving average of the student model parameters. The update formula is expressed:
\begin{equation}
    \Theta_{t}^{te}=\alpha\Theta_{t-1}^{te}+(1-\alpha)\Theta_{t}^{st}
\end{equation}
where $\alpha$ is the smoothing coefficient simply set at $0.999$. This strategy alleviates the over-fitting risk of noisy pseudo labels in the beginning of the training process. 

\subsection{Curriculum Learning}
TFDA implements the curriculum learning strategy \cite{Bengio2009CurriculumL,Karim2023CSFDAAC} focusing on easy samples first before turning into hard samples, thus avoiding early memorization of noisy pseudo labels. Since a model is confident to the reliable group $\mathbb{R}$, their predictions are likely to be correct. Hence, the priority is to learn the reliable group first where the non-reliable group $\mathbb{\tilde{R}}$ is restricted in the beginning of the training process. This is achieved by setting the tradeoff coefficient:
\begin{equation}
    \mu_{r}^{j}=\mu_{r}^{j-1}(1-\alpha\exp{(-\frac{1}{d^{j}})})
\end{equation}
where $d^{j}=\frac{\tau_u}{\tau_c}$ denotes the difficulty score. $\alpha,\mu_{r}^{0}$ are empirically set to $0.005$ and $1$. This setting implies to prevent the non-reliable group to be learned in the beginning of the training process. The learning process of the non-reliable group commences as the training process progresses and the overall reliability improves. Besides, the dynamics of $\mu_{r}$ is determined directly from the difficulty score meaning that the change is minimal if the current batch of samples at the epoch $j$ is difficult. $\mu_{c}, \mu_{cons}, \mu_{u}$ exponentially decay. 
\begin{equation}
    \mu_{c}^{j}=\mu_{c}^{j-1}\exp{(-\beta)} 
\end{equation}
\begin{equation}
    \mu_{cons}^{j}=\mu_{cons}^{j-1}\exp{(-\beta)} 
\end{equation}
\begin{equation}
    \mu_{u}^{j}=\mu_{u}^{j-1}\exp{(-\beta)} 
\end{equation}
where $\mu_{c}^{0},\mu_{cons}^{0},\mu_{u}^{0},\beta$ are respectively set as $0.5, 0.5, 0.5, 1e-4$. This setting allows them to be active in the beginning of the training process but to be slow in the end of the training process. That is, fine-grained pseudo labels are supposed to be achieved in the end of the training process. 

\subsection{Theoretical Analysis}    
Given $\mathit{P^{S}}, \mathit{P^{T}}$ as the distribution of labeled source and unlabeled target examples respectively over source $\mathcal{X}^S$ and target $\mathcal{X}^T$ input spaces, where $\mathit{P^S \neq P^T}$. Let $\widehat{P} := \left\{ x_1,\cdots,x_n \right\} \subset \mathcal{X}$ denote $n$ uniformly distributed unlabeled training data from $P^T$. We consider $F:\mathcal{X}\rightarrow \mathcal{Z}$ as a continous logits output by a feature extractor, and $G:\mathcal{X}\rightarrow[K]$ the discrete labels induced by $F:G(x):= \textnormal{argmax}_i F(x)_i$. $K$ is the number of classes given by the ground truth $G^*(x)$ for $G^*:\mathcal{X}\rightarrow[K]$, $F^*:G(x)$ as a ground truth feature extractor, and $\mathcal{Q}_i:=\left\{ x:G^*(x)=i \right\}$ as the set of examples with ground-truth label $i$. Let define $\mathcal{A}(x)$ as the augmentation function of input $x$, where $\mathcal{A}(x):= \left\{ x' |\exists T \in \mathcal{T}_{wa} \textnormal{ such that} \left\| x'-T(x) \right\| \leq r \right\}$, $\mathcal{T}_{wa}$ denote some set of augmented data, $\mathcal{N}(x):=\{ x'|\mathcal{A}(x) \cap  \mathcal{A}(x')\neq \emptyset  \}$ as the neighborhoods of $x$, and $\mathcal{N}(S):=\cup_{x\in S} \mathcal{N}(x)$ as the neighborhood set of $S$. Based on self-learning expansion assumption \cite{Wei2020TheoreticalAO}, let define ($a$,$b$)-multiplicative-expansion:
\begin{equation}\label{def:exp}
    P_i(\mathcal{N}(S))\ge \textnormal{min}\left\{ bP_i(S),1 \right\}
\end{equation}
for $P_i(S)\leq 1/2, b > 1$, where $P_i$ be the distribution of subset $S$ on class $i$ data and $b$ represents an expansion factor that corresponds to the strength of the data augmentation. Let also define ($c$,$\rho$)-constant-expansion:
\begin{equation}\label{def:exp-con}
    P(\mathcal{N^*}(S)\backslash S) \ge \textnormal{min}\left\{ \rho,P(S) \right\}
\end{equation}
for $P(S)\ge c$ and $P(S \cap \mathcal{Q}_i) \leq P(\mathcal{Q}_i)/2$ for all $i$, where $\mathcal{N^*}(S)$ is the neighborhood of $S$ with neighbors restricted to the same class: $\mathcal{N^*}(S):=\cup_{i\in [K]}(\mathcal{N}(S\cap \mathcal{Q}_i)\cap \mathcal{Q}_i)$.
Our algorithms leverage expansion by using cross-entropy and transductive label propagation loss (\ref{eq:self-learning}) as consistency regularization to
encourage predictions of a classifier $G$ to be consistent under different transformations of the data. Furthermore, let consider side information \cite{ge2023provableadvantageunsupervisedpretraining} $s \in S$ which can be accessed from $x \in \mathcal{X}$. In the case of contrastive learning, given $(x, x^o)\in \mathcal{X}^2, s:=\mathbb{1}(x=x^o)$ where side information indicates whether the pair should be considered as positive or negative. We follow the self-learning expansion and separation assumptions as in \cite{Wei2020TheoreticalAO} and $\mathcal{_K}^{-1}$-informative condition assumption as in \cite{ge2023provableadvantageunsupervisedpretraining} as detailed below:
\begin{assumption}\label{as:1}(multiplicative-expansion)
Assume P satisfies (1/2,b)-expansion on $\mathcal{X}$ for $b > 1$, ground-truth classes $G^*$ are separated, and the consistency regularization loss as $\mathcal{R_A}(G):=\mathbb{E}_P\left[ \textnormal{\textbf{1}}(\exists x' \in \mathcal{A}(x) \textnormal{ such that } F(x')\neq F(x)) \right]$. 
\end{assumption}
\begin{assumption}\label{as:2}(separation) Assume P is $\mathcal{A}$-separated with probability $1-\mu$ by ground-truth classifier $G^*$, as follows: $\mathcal{R_A}(G^*) \leq \mu$, $\mu$ represents a small negligible value.
\end{assumption}
\begin{assumption}\label{as:3}
(constant-expansion)
    Assume P satisfies ($c$,$\rho$)-constant-expansion for some c, $\mathcal{R_A}(G) < \rho$ and $\underset{i}{\textnormal{min}} P(\left\{ x:G(x)=i \right\})>2\textnormal{ max}\left\{ c,\mathcal{R_A}(G) \right\}$, there exist permutation $\pi:[K]\rightarrow[K]$ satisfies 
    
    $P(\left\{ x:\pi(G(x))\neq G^*(x) \right\})\le \textnormal{max}\left\{ c,\mathcal{R_A}(G) \right\}+\mathcal{R_A}(G)$ 
\end{assumption}
\begin{assumption}
    ($\mathcal{_K}^{-1}$-informative condition) Assume that feature-extractor model $f_{\theta}$ is $\mathcal{_K}^{-1}$-informative w.r.t. the true model $f_{\theta}^*, g_{\phi}^*$ if for any $f_{\theta}\in \Theta$ and $x\in \mathcal{X}$ s.t.
    $\mathcal{L}_{CE}(g_{\phi}^*\circ f_{\theta}(x),y)\le _\mathcal{K}\mathbb{E}_{x^o}[\mathcal{L}_{cl}(f_\theta,f_{\theta}^*,(x,x^o),s)]$
\end{assumption}
\begin{theorem}\label{th:1}
Suppose that Assumptions (\ref{as:1}), (\ref{as:2}) and (\ref{as:3}) hold for some b, $\mu$ such that $\textnormal{min}_{y\in [K]}P(\left\{ x:G^*(x)=y \right\}) > max \left\{ 2/(b-1),2 \right\}\mu$. Then any minimizer $\widehat{G}$ of
\begin{multline}
    \underset{G}{\textnormal{min }}\mathcal{R_A}(G) \textnormal{  subject to  } \underset{y\in [K]}{\textnormal{min }} \mathbb{E}_P\left[ \textnormal{\textbf{1}}(G(x)=y) \right] > \\ \textnormal{max}\left\{ \frac{2}{b-1},2 \right\} \mathcal{R_A}(G)
\end{multline}
satisfies
\begin{equation}
\textnormal{Err}_{\textnormal{unsup}}(\widehat{G})\leq \textnormal{max}\left\{ \frac{2}{b-1},2 \right\}\mu
\end{equation}

\end{theorem}
\begin{theorem}\label{th:2}
    Let $\widehat{f}_{\theta}$ be the minimizer of equation (\ref{eq:cl}). Then with probability at least $1-\delta$, we have 
    \begin{equation}
    d_{TV}(\mathbb{P}_{\widehat{f}_{\theta}}(x,s),\mathbb{P}_{f_{\theta}^*}(x,s))\le 3\sqrt{\frac{1}{n^2}log\frac{N_{[]}(P_{\mathcal{X}\times \mathcal{S}}(\Theta),\frac{1}{n^2})}{\delta}}
    \end{equation}
    where $P_{\mathcal{X \times \mathcal{S}}}(\Theta) = \left\{ \mathbb{P}_{f_{\theta}} (x,s)|f_{\theta}\in \Theta\right\}, x=(x_i,x_j)$, $s$ indicates whether $x_i,x_j$ is the paired data, and $N_{[]}(.,.)$ denotes the bracket number.
\end{theorem}

Theorem (\ref{th:1}) prove that our self-learning strategy leveraging entropy minimization and transductive label propagation loss has an upper bound on the target error, while theorem (\ref{th:2}) prove that with contrastive learning the total variation distance between the best feature representation $f_{\theta}^*$
and the learned feature representation $\hat{f_{\theta}}$ can be bounded by the bracket number
of the possible distribution space $P_{\mathcal{X}\times \mathcal{S}}(\Theta)$.

\subsection{Complexity Analysis}
Suppose that $E,N,M$ respectively stand for the number of epochs, the number of target-domain samples and the size of memory bank where $M<N$, the complexity of TFDA is approximated to be $\approx O(E.M.N)$. Since the number of epochs and the size of memory bank are bounded and usually small, the complexity of TFDA is $\approx O(N)$. Detailed derivations of the algorithm's complexity and pseudo-code can be found in the appendix.


\section{Experiments}
\subsection{Datasets}
Our method, TFDA, is evaluated with three datasets of different application domains, human activity recognition, machine fault diagnosis and sleep stage classification. Table \ref{tab:data} summarizes the key characteristics of the three datasets. 
\newline\noindent\textbf{UCIHAR dataset} describes an activity recognition task using three sensors providing three dimensional reading of body movements resulting in 9 channels per sample. We follow the same configuration as per \cite{Ragab2023SourceFreeDA} where our experiments are conducted in five cross-users experiments. That is, a model is trained with one user and subsequently evaluated with another user to reflect the cross-domain scenario. 
\newline\noindent\textbf{Sleep Stage Classification (SSC) dataset} is a classification dataset based on electroencephalography (EEG) signal classified into five classes. As with \cite{Ragab2023SourceFreeDA}, the sleep-EDF dataset \cite{Goldberger2000PhysionetCO} is used. As with \cite{Eldele2021AnAD}, a single channel, namely Fpz-Cz, is selected and 5 cross-domain experiments are performed from 10 subjects. 
\newline\noindent\textbf{MFD dataset} constitutes a bearing fault diagnosis dataset \cite{Lessmeier2016ConditionMO} of the University of Paderborn where the fault condition is detected via the vibration signal. This dataset possesses four different working conditions where each working condition represents one domain. We follow \cite{Ragab2023SourceFreeDA} where five cross-conditions experiments are selected.  

\begin{table}[]
\caption{Dataset characteristics (Ch: \# channels, K: \# classes, S: sample length)}\label{tab:data}
\centering
\begin{tabular}{l|ccc|cc}
\hline
\multicolumn{1}{c|}{Dataset} & Ch & K & S    & \# Training & \# Testing \\ \hline
MFD                          & 1 & 3 & 5120 & 7312             & 3604            \\
UCIHAR                       & 9 & 6 & 128  & 2300             & 990             \\
SSC                          & 1 & 5 & 3000 & 14280            & 6130            \\
 \hline
\end{tabular}
\end{table}

\subsection{Hyperparameters}
We apply the same network architecture as per \cite{Eldele2021TimeSeriesRL,Ragab2023SourceFreeDA} to ensure fair comparisons. That is, each encoder, i.e., time and frequency, is configured as the three-layers one-dimensional convolutional neural network with the filter size of $64, 128, 128$ respectively. Each convolutional layer is accompanied by the ReLU layer and the batch normalization layer. To further ensure fair and valid comparisons, the benchmark implementations of \cite{Liang2020DoWR,Yang2021ExploitingTI,Yang2022AttractingAD,Ragab2023SourceFreeDA} are followed. Our algorithm is run with a batch size of $32$, learning rate of $1e-6$, and dropout ratio of $0.5$. For evaluation metric, we apply the macro F1 score offering decent performance evaluation under the class imbalanced situations. Our algorithm is run 3 times consecutively with distinct random seeds. We report the average accuracy across the three runs with standard deviations. 

\subsection{Benchmark Algorithms}
TFDA is compared against conventional UDA algorithms having access to the source-domain samples: DDC \cite{Tzeng2014DeepDC}, DCORAL \cite{Sun2016CorrelationAF}, HoMM \cite{Chen2019HoMMHM}, MMDA \cite{Rahman2019OnMD}, DANN \cite{Ganin2014UnsupervisedDA}, CDAN \cite{Long2017ConditionalAD}, CoDATS \cite{Wilson2020MultiSourceDD}, AdvSKM \cite{Liu2021AdversarialSK}. In addition, we also compare our method with source-free domain adaptation methods: SHOT \cite{Liang2020DoWR}, NRC \cite{Yang2021ExploitingTI}, AaD \cite{Yang2022AttractingAD}, MAPU \cite{Ragab2023SourceFreeDA}. All algorithms are run under the same computational resources, NVIDIA A5000 GPU with 48 GB RAM, using their official implementations under the same backbone networks and training setting. TFDA is implemented under the pytorch library where its source codes are publicly shared in \url{https://github.com/furqon3009/TFDA}.

\subsection{Numerical Results}
Table \ref{tab:har} reports numerical results of compared algorithms in the UCIHAR dataset. The advantage of TFDA is clear here where it produces the highest average accuracy across five cross-domain cases ahead of MAPU in the second place with over $0.5\%$ margin. In particular, TFDA delivers the highest F1 score in the two cases $2\rightarrow11$ and $9\rightarrow18$. The numerical results of all consolidated algorithms in the SSC dataset is presented in Table \ref{tab:eeg} where TFDA produces the best result on average across five cross-domain cases with a marginal gap around $~0.2\%$ to MMDA in the second place. TFDA generates the best results in two cases $16\rightarrow1$ and $7\rightarrow18$. Note that MMDA fully exploits the source domain samples while TFDA is a fully source-free method. Table \ref{tab:mfd} exhibits the numerical results of all consolidated algorithms in the MFD dataset. TFDA attains the highest average performance compared to other algorithms with around $~1.1\%$ difference to MaPU in the second place. Our method produces the highest F1 score in one case, $1\rightarrow0$ and the second best F1 score in one case, $0\rightarrow1$. Although some methods delivers higher performances than ours in some cases, their performances drop in other cases worsening their average performances. 

\begin{table*}[]
\caption{Five UCIHAR cross-domain scenarios results in terms of MF1 score. SF means Source-Free.}\label{tab:har}
\centering
\begin{tabular}{l|c|ccccc|c}
\hline
\multicolumn{1}{c|}{Method} & SF & 2 $\rightarrow$ 11 & 6 $\rightarrow$ 23 & 7 $\rightarrow$ 13 & 9 $\rightarrow$ 18 & 12 $\rightarrow$ 16       & AVG            \\ \hline
AdvSKM        & \xmark  & 65.74±2.69          & 79.63±8.52          & 88.89±3.12          & 53.25±5.19          & 60.52±1.99          & 74.67          \\
CDAN          & \xmark  & 98.19±1.57          & 96.73±0.00          & \underline{ 93.33±0.00}    & 71.3±14.64          & 61.20±3.27          & 86.79          \\
CoDATS        & \xmark  & 86.65±4.28          & 92.08±4.39          & 92.61±0.51          & \underline{ 80.51±8.47}    & 61.03±2.33          & 85.47          \\
DANN          & \xmark  & 98.09±1.68          & 85.6±15.71          & \underline{ 93.33±0.00}    & 70.7±11.36          & 62.08±1.69          & 84.97          \\
DCoral        & \xmark  & 67.2±13.67          & 89.66±2.54          & 90.46±2.96          & 54.38±9.69          & 64.58±8.72          & 77.71          \\
DDC           & \xmark  & 60.0±13.32          & 88.55±1.42          & 77.29±2.11          & 61.41±5.80          & 66.77±8.46          & 75.67          \\
HoMM          & \xmark  & 83.54±2.99          & 94.97±2.49          & 91.41±1.33          & 71.25±4.42          & 63.45±2.07          & 84.1           \\
MMDA          & \xmark  & 72.91±2.78          & 91.14±0.46          & 90.61±2.00          & 62.62±2.63          & \underline{ 74.64±2.88}    & 81.4           \\ \hline
AaD           & \cmark  & \underline{ 98.51±2.58}    & \underline{ 98.07±1.71}    & 89.41±2.86          & 68.33±11.9          & 66.15±6.15          & 84.09          \\
MAPU          & \cmark  & \textbf{100.0±0.00} & \textbf{98.91±1.89} & \textbf{99.29±1.22} & 78.46±1.44          & \underline{ 71.24±4.75}    & \underline{ 89.57}    \\
NRC           & \cmark  & 97.02±2.82          & 96.41±1.33          & 89.13±0.54          & 63.10±4.84          & 72.18±0.59          & 83.57          \\
SHOT          & \cmark  & \textbf{100.0±0.00} & \textbf{98.91±1.89} & 93.01±0.57          & 70.19±8.99          & 70.76±6.22          & 86.57          \\ \hline
\textbf{TFDA} & \cmark  & \textbf{100.0±0.00} & 89.27±2.38          & 92.66±7.01          & \textbf{91.80±3.34} & \textbf{76.73±3.39} & \textbf{90.09} \\ \hline
\end{tabular}
\end{table*}

\begin{table*}[]
\caption{Five SSC cross-domain scenarios results in terms of MF1 score. SF means Source-Free.}\label{tab:eeg}
\centering
\begin{tabular}{l|c|ccccc|c}
\hline
\multicolumn{1}{c|}{Method} & SF & 0 $\rightarrow$ 11  & 7 $\rightarrow$ 18   & 9 $\rightarrow$ 14   & 12 $\rightarrow$ 5   & 16 $\rightarrow$ 1   &     AVG            \\ \hline
AdvSKM        & \xmark  & \textbf{55.19±4.19} & 67.58±3.64          & 64.76±3.00          & 55.73±1.42          & 57.83±1.42          & 60.21         \\
CDAN          & \xmark  & 39.38±3.28          & 67.61±3.55          & 64.18±6.37          & 64.43±1.17          & 59.65±4.96          & 59.04         \\
CoDATS        & \xmark  & 46.28±5.99          & 66.06±2.48          & 63.51±6.92          & 52.54±5.94          & \underline{ 63.84±3.36}    & 58.44         \\
DANN          & \xmark  & 44.13±5.84          & 69.54±3.00          & 64.29±1.08          & \underline{ 64.65±1.83}    & 58.68±3.29          & 60.26         \\
DCoral        & \xmark  & 53.76±1.89          & 67.49±1.50          & 63.50±1.36          & 55.35±2.64          & 55.50±1.74          & 59.12         \\
DDC           & \xmark  & \underline{ 54.17±1.79}    & 67.46±1.45          & 63.57±1.43          & 55.43±2.75          & 55.47±1.72          & 59.22         \\
HoMM          & \xmark  & 53.37±2.47          & 67.50±1.50          & 63.49±1.14          & 55.46±2.71          & 55.51±1.79          & 59.06         \\
MMDA          & \xmark  & 43.23±4.31          & \underline{ 70.95±0.82}    & 71.04±2.39          & \textbf{65.11±1.08} & 62.92±0.96          & \underline{ 62.79}   \\ \hline
AaD           & \cmark  & 44.04±2.18          & 67.35±1.48          & 65.27±1.69          & 61.84±1.74          & 57.04±2.03          & 59.11         \\
MAPU          & \cmark  & 43.42±7.38          & 65.04±0.95          & \textbf{73.55±0.35} & 63.58±1.56          & 62.95±0.39          & 61.71         \\
NRC           & \cmark  & 47.55±1.72          & 66.18±0.25          & 58.52±0.66          & 59.87±2.48          & 52.09±1.89          & 56.84         \\
SHOT          & \cmark  & 50.78±1.90          & 69.74±1.22          & 69.93±0.46          & 62.11±1.62          & 59.07±2.14          & 62.33         \\ \hline
\textbf{TFDA} & \cmark  & 45.51±9.51          & \textbf{71.81±3.03} & \underline{ 71.64±1.63}    & 60.79±3.49          & \textbf{65.75±1.97} & \textbf{62.9}    \\ \hline
\end{tabular}
\end{table*}

\begin{table*}[]
\caption{Five MFD cross-domain scenarios results in terms of MF1 score. SF means Source-Free.}\label{tab:mfd}
\centering
\begin{tabular}{l|c|ccccc|c}
\hline
\multicolumn{1}{c|}{Method} & SF & 0 $\rightarrow$ 1    & 1 $\rightarrow$ 0    & 1 $\rightarrow$ 2    & 2 $\rightarrow$ 3    & 3 $\rightarrow$ 1          & AVG                  \\ \hline
AdvSKM        & \xmark  & \textbf{100.0±0.00} & 98.85±0.93          & 83.10±2.19          & 43.81±6.29          & 76.64±4.82          & 80.48          \\
CDAN          & \xmark  & \textbf{100.0±0.00} & 99.7±0.45           & 85.96±0.90          & 52.39±0.49          & 84.97±0.62          & 84.6           \\
CoDATS        & \xmark  & 99.92±0.14          & 99.21±0.79          & 89.05±4.73          & 49.92±13.7          & 67.42±13.3          & 81.1           \\
DANN          & \xmark  & \textbf{100.0±0.00} & \underline{ 99.95±0.09}    & 84.19±2.10          & 51.52±0.38          & 83.44±1.72          & 83.82          \\
DCoral        & \xmark  & 97.73±3.93          & 98.01±0.67          & 82.71±0.76          & 40.83±5.01          & 79.03±8.83          & 79.66          \\
DDC           & \xmark  & \textbf{100.0±0.00} & 96.34±3.07          & \underline{ 89.34±2.16}    & 48.91±6.24          & 74.50±5.56          & 81.82          \\
HoMM          & \xmark  & 96.28±6.45          & 98.61±0.08          & 84.28±1.32          & 42.31±5.90          & 80.80±2.46          & 80.46          \\
MMDA          & \xmark  & \textbf{100.0±0.00} & \textbf{100.0±0.00} & \textbf{94.07±2.72} & 49.35±5.02          & 82.44±4.47          & 85.17          \\ \hline
AaD           & \cmark  & 87.45±11.7          & 90.07±7.02          & 78.31±2.26          & 75.33±4.65          & 71.72±3.96          & 80.58          \\
MAPU          & \cmark  & 99.43±0.99          & 98.81±1.79          & 85.75±7.34          & \underline{ 77.24±0.31}    & \textbf{97.25±2.46} & \underline{ 91.69}    \\
NRC           & \cmark  & 71.48±4.59          & 78.04±11.3          & 69.23±0.75          & 74.88±8.81          & 73.99±1.36          & 73.52          \\
SHOT          & \cmark  & \underline{ 99.95±0.05}    & 99.48±0.31          & 80.70±1.49          & 57.00±0.09          & 41.99±2.78          & 75.82          \\ \hline
\textbf{TFDA} & \cmark  & 99.54±0.73          & 99.32±0.21          & 82.75±4.06          & \textbf{88.38±6.68} & \underline{ 93.86±1.07}    & \textbf{92.77} \\ \hline
\end{tabular}
\end{table*}

\begin{figure}
    \centering
    \includegraphics[width=0.9\linewidth]{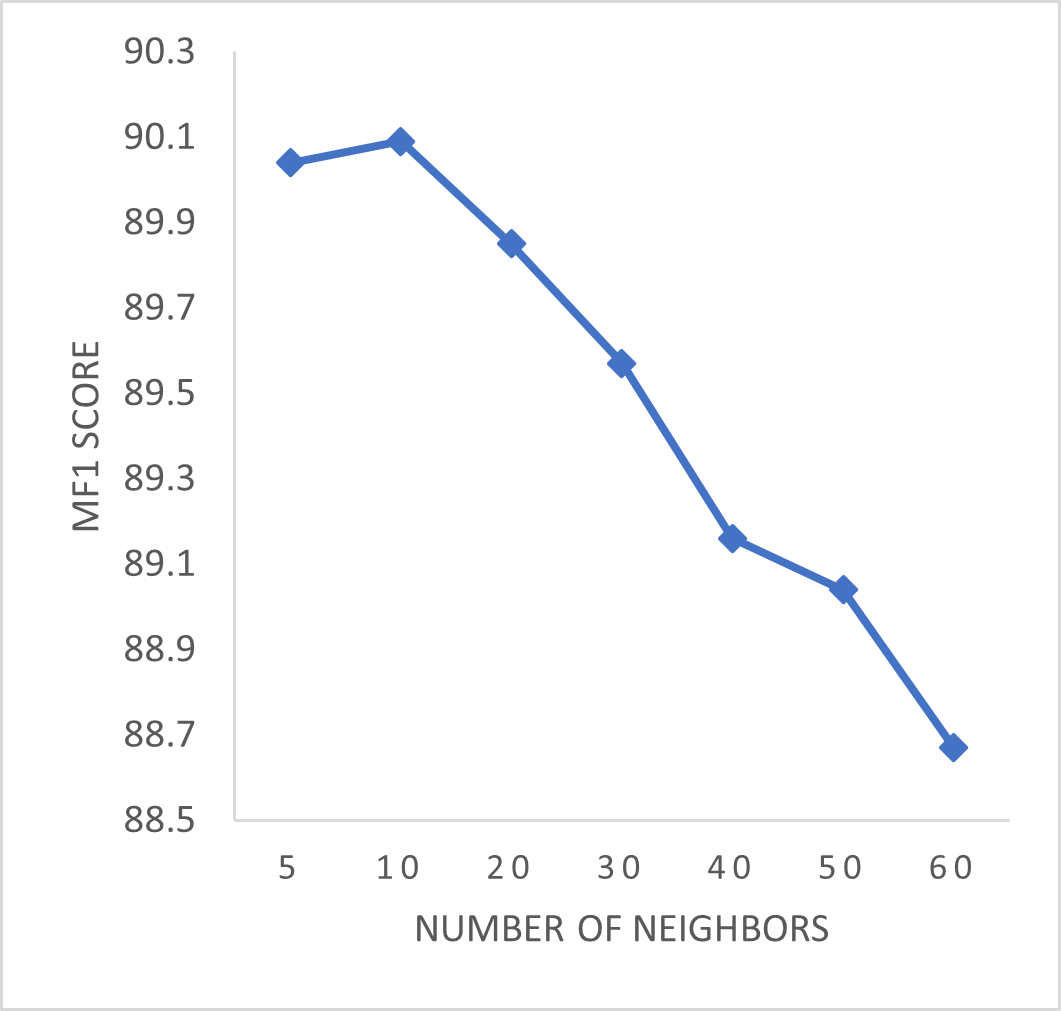}
    \caption{MF1 score results with different number of neighbors}
    \label{fig:numNgbr}
\end{figure}

\begin{table}[]
\caption{Ablation study}\label{tab:ablation}
\centering
\begin{tabular}{lcl}
\cline{1-2}
\multicolumn{1}{c}{Selection   Strategy} & MF1 Score      &  \\ \cline{1-2}
Ours w/o curriculum learning             & 86.76          &  \\
Ours w/o contrastive learning $\mathcal{L}_{CL}$   & 55.92          &  \\
Ours w/o uncertainty reduction $\mathcal{L}_{ul}$           & 86.70          &  \\
Ours w/o label propagation $\mathcal{L}_{lp}$              & 88.92          &  \\
Ours w/o self distillation $\mathcal{L}_{cons}$              & 86.96          &  \\ \cline{1-2}
\textbf{Ours}                            & \textbf{90.09} &  \\ \cline{1-2}
\end{tabular}
\end{table}

\subsection{Ablation Study}
Ablation study is performed to check the contribution of each learning component of TFDA whether it contributes positively to the overall learning performance of TFDA. It is done using the UCIHAR dataset where Table \ref{tab:ablation} reports the average F1 score of the five cross-domain cases. It is seen that the absence of curriculum learning affects to the average F1 score of TFDA where reduction by about $4\%$ is perceived. The absence of curriculum learning results in early memorization of noisy pseudo labels. The same case applies without the contrastive learning module where drops of average F1 score by about $35\%$ occurs. Note that this situation implies the removal of the overall contrastive learning $\mathcal{L}_{CL}$, i.e., the absence of the time contrastive learning  $\mathcal{L}_{cl}$, the frequency contrastive learning $\mathcal{L}_{cl}^{f}$ and the time-frequency contrastive learning $\mathcal{L}_{cl}^{tf}$. This situation ensues because of the neighborhood pseudo-labelling mechanism which relies on the assumption that similar samples should share the same pseudo labels. This assumption holds with the contrastive learning strategy which pulls similar samples together while pushing apart dissimilar samples. In addition, the consistency in each domain and across time and frequency domains is not ensured with the absence of the contrastive learning framework. The absence of uncertainty reduction strategy contributes to $3\%$ performance deterioration. This module is vital to minimize the prediction uncertainties due to the domain shifts problem. The label propagation strategy affects similarly as other learning components where its deactivation brings down the average F1 score by about $1\%$. This configuration implies the full use of class balanced cross-entropy in learning all data samples. Last but not least, the deactivation of the self-distillation drops the performance by about $4\%$. The self-distillation technique is vital because it aligns the temporal and frequency representations, i.e., the key step in domain adaptation. 

\subsection{Analysis of Neighborhood Size}
This section is meant to study the influence of neighborhood size to the performance of TFDA. Fig. \ref{fig:numNgbr} shows the average F1 score of TFDA against varying neighborhood sizes. It is seen that the performance of TFDA improves as the increase of neighborhood size. This fact confirms the advantage of neighborhood pseudo labelling strategy where pseudo labels are induced by a group of similar samples. Nevertheless, the performance of TFDA significantly degrades with over $10$ neighbors because of noisy pseudo labels. That is, distinct samples are included when aggregating the pseudo labels. The number of neighbors is fixed to $10$ to all cases. 

\begin{figure}[h]
    \centering
    \includegraphics[width=0.9\linewidth]{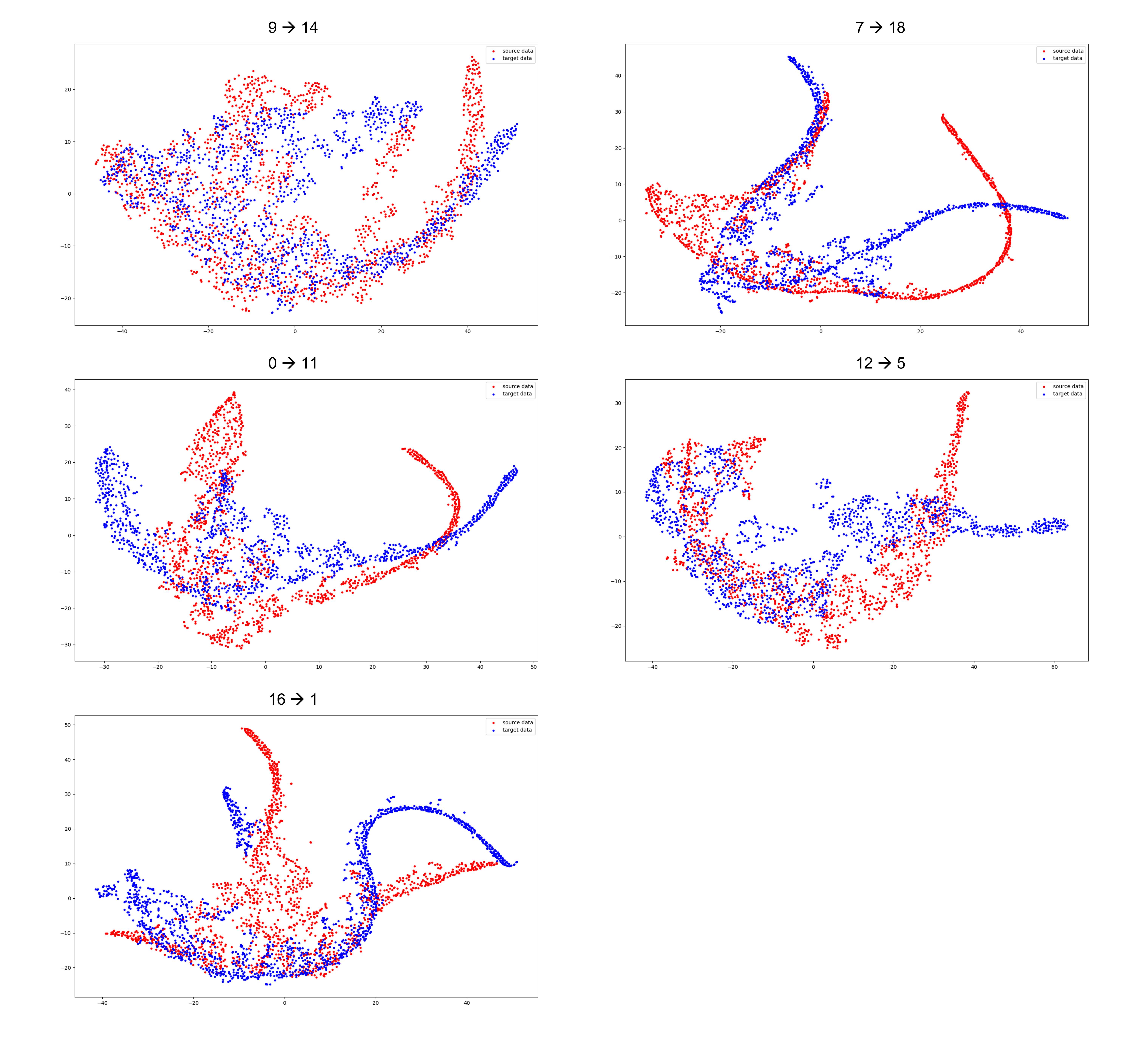}
    \caption{T-SNE of SSC Dataset}
    \label{fig:tsne}
\end{figure}

\subsection{T-SNE Analysis}
Fig. \ref{fig:tsne} depicts the t-sne plots of TFDA in the SSC dataset. It is seen that TFDA delivers decent embedding qualities where the source and target domain samples are mapped closely in the embedding space with the absence of any source domain samples during the domain adaptation phase. 

\subsection{Analysis of Execution Times}
Table \ref{tab:runtime} tabulates execution times of TFDA across all domain adaptation cases. It is seen that TFDA's execution times are relatively fast to be run under our machine, the NVIDIA A5000 GPU with 48 GB of RAM. This facet demonstrates its feasibility to be deployed in the practical source-free time-series domain adaptation cases.

\begin{table}[]
\footnotesize
\caption{TFDA Execution Times (in seconds)}\label{tab:runtime}
\centering
\begin{tabular}{clllll}
\hline
\multicolumn{6}{c}{MFD   Dataset}                                                                                                                                                                                              \\ \hline
0 $\rightarrow$ 1           & \multicolumn{1}{c}{1 $\rightarrow$ 2}  & \multicolumn{1}{c}{3 $\rightarrow$ 1}  & \multicolumn{1}{c}{1 $\rightarrow$ 0}  & \multicolumn{1}{c}{2 $\rightarrow$ 3}   & \multicolumn{1}{c}{Average} \\
\multicolumn{1}{l}{610.118} & 575.761                                & 515.848                                & 547.714                                & 507.458                                 & 551.3798                    \\ \hline
\multicolumn{6}{c}{HAR Dataset}                                                                                                                                                                                                \\ \hline
2 $\rightarrow$ 11          & \multicolumn{1}{c}{6 $\rightarrow$ 23} & \multicolumn{1}{c}{7 $\rightarrow$ 13} & \multicolumn{1}{c}{9 $\rightarrow$ 18} & \multicolumn{1}{c}{12 $\rightarrow$ 16} & \multicolumn{1}{c}{Average} \\
\multicolumn{1}{l}{176.055} & 227.632                                & 195.587                                & 197.113                                & 217.342                                 & 202.7458                    \\ \hline
\multicolumn{6}{c}{EEG Dataset}                                                                                                                                                                                                \\ \hline
0 $\rightarrow$ 11          & \multicolumn{1}{c}{12 $\rightarrow$ 5} & \multicolumn{1}{c}{7 $\rightarrow$ 18} & \multicolumn{1}{c}{16 $\rightarrow$ 1} & \multicolumn{1}{c}{9 $\rightarrow$ 14}  & \multicolumn{1}{c}{Average} \\
\multicolumn{1}{l}{330.157} & 354.417                                & 361.845                                & 404.395                                & 340.855                                 & 358.3338                    \\ \hline
\end{tabular}
\end{table}

\section{Conclusion}
This paper presents Time Frequncy Domain Adaptation (TFDA) to overcome the absence of source-domain samples in time-series unsupervised domain adaptation. TFDA is built upon the dual-branch network architecture processing the temporal and frequency features separately while final predictions are drawn from the confidence score. It relies on neighborhood pseudo labelling aided by the contrastive learning strategy in both time and frequency domains to pull similar samples while pushing dissimilar samples. The domain adaptation step is achieved by enforcing consistencies between the time and frequency features. In addition, the uncertainty reduction strategy is implemented to alleviate the prediction's uncertainties due to the domain shifts and the curriculum learning strategy is designed to prevent early memorization of noisy pseudo labels. That is, the curriculum learning strategy prioritizes in learning reliable samples while avoiding those of unreliable samples. Rigorous experiments are performed to numerically validate the efficacy of TFDA. It is demonstrated that TFDA delivers the most encouraging results across $15$ cross-domain scenarios using the three datasets with noticeable margins to its competitors. Our ablation study further substantiates the efficacy of each learning component while the analysis of neighborhood size exhibits the advantage of the neighborhood pseudo-labelling strategy. Notwithstanding that TFDA delivers satisfactory performances compared to prior arts, TFDA assumes that the source model can be shared when performing domain adaptations. The access of source model is prohibited in some applications due to the privacy constraints. Our future work is directed to study the topic of black box domain adaptation.

\appendix
\renewcommand{\theequation}{A.\arabic{equation}}
\setcounter{equation}{0}
\section{Detailed Theoretical Analysis}
Given $\mathit{P_{S}}, \mathit{P_{T}}$ as the distribution of labeled source and unlabeled target examples respectively over source $\mathcal{X}^S$ and target $\mathcal{X}^T$ input spaces, where $\mathit{P_S \neq P_T}$. Let $\widehat{P} := \left\{ x_1,\cdots,x_n \right\} \subset \mathcal{X}$ denote $n$ uniformly distributed unlabeled training data from $P_T$. We consider $F:\mathcal{X}\rightarrow \mathcal{Z}$ as a continous logits output by a feature extractor, and $G:\mathcal{X}\rightarrow[K]$ the discrete labels induced by $F:G(x):= \textnormal{argmax}_i F(x)_i$. $K$ is the number of classes given by the ground truth $G^*(x)$ for $G^*:\mathcal{X}\rightarrow[K]$ and $\mathcal{Q}_i:=\left\{ x:G^*(x)=i \right\}$ as the set of examples with ground-truth label $i$. Let define $\mathcal{A}(x)$ as the augmentation function of input $x$, where $\mathcal{A}(x):= \{ x' |\exists T \in \mathcal{T}_{wa} \textnormal{ such that} \left\| x'-T(x) \right\| \leq r \}$, $\mathcal{T}_{wa}$ denote some set of augmented data, $\mathcal{N}(x):=\left\{ x'|\mathcal{A}(x)\cap \mathcal{A}(x')\neq \emptyset  \right\}$ as the neighborhoods of $x$, and $\mathcal{N}(S):=\cup_{x\in S} \mathcal{N}(x)$ as the neighborhood set of $S$. Based on self-learning expansion assumption \cite{Wei2020TheoreticalAO}, $P_i(\mathcal{N}(S)) \ge bP_i(S)$ for $P_i(S)\leq 1/2, b > 1$, where $P_i$ be the distribution of subset $S$ on class $i$ data and $b$ represents an expansion factor that corresponds to the strength of the data augmentation.
We follow the self-learning expansion and separation assumptions adapted from \cite{Wei2020TheoreticalAO}.

\subsection{Proofs for Theorem 1}
To prove Theorem (\ref{th:1}), we convert (a,b)-multiplicative-expansion (\ref{def:exp}) to (c,$\rho$)-constant-expansion using following lemma

\begin{lemma}\label{le:2}
Suppose P satisfy (1/2,b)-multiplicative-expansion on $\mathcal{X}$. Then for any choice of $\rho$>0, P satisfies $\left( \frac{\rho}{b-1},\rho \right)$-constant-expansion
\end{lemma}
$\textit{Proof}$. Consider any $\mathit{S}$ s.t. $P(S\cap \mathcal{Q}_i)\le P(\mathcal{Q}_i)/2$ for all $i \in [K]$ and $P(S)>c$. Define $S_i:=S\cap \mathcal{Q}_i$. When $b\ge 2$ we have multiplicative-expansion
\begin{align}
    \begin{split}\label{le:3}
        P(\mathcal{N^*}(S)\backslash S)&\ge \sum_{i}P(\mathcal{N^*}(S_i))-P(S_i)
    \end{split}\\
    \begin{split}
        &\ge \sum_{i}min\left\{ bP(S_i),P(\mathcal{Q}_i) \right\}-P(S_i)
    \end{split}
\end{align}
    
    substituting  $P(S_i) \leq P(\mathcal{Q}_i)/2$ and $b \ge 2$ then we have
    
\begin{align}
    \begin{split}
        P(\mathcal{N^*}(S)\backslash S)&\ge \sum_{i}min\left\{ 2P(S_i),2P(S_i) \right\}-P(S_i)
    \end{split}\\
    \begin{split}
        &\ge \sum_{i}P(S_i)
    \end{split}
\end{align}

Thus, we obtain constant expansion.

Considering the case where $1 \leq b < 2$, again by multiplicative expansion as in (\ref{le:3}), substituting  $P(\mathcal{Q}_i) \ge 2P(S_i)$ and $b < 2$ we have
\begin{align}
    \begin{split}
        P(\mathcal{N^*}(S)\backslash S)&\ge \sum_{i}min\left\{ bP(S_i),2P(S_i) \right\}-P(S_i)
    \end{split}\\
    \begin{split}
        &\ge \sum_{i} bP(S_i)-P(S_i) = \sum_{i} (b-1)P(S_i)
    \end{split}\\
    \begin{split}
        &\ge (b-1)c=\rho
    \end{split}
\end{align}

The following lemma states an accuracy guarantee for the setting with multiplicative expansion.

\begin{lemma}
    Suppose Assumption (\ref{as:1}) holds for some b > 1. If classifier G satisfies
    \begin{align}
        \underset{i}{\textnormal{min}}\mathbb{E}_P\left[ \textnormal{\textbf{1}} (G(x)=i)\right] > \textnormal{max}\left\{ \frac{2}{b-1},2 \right\}\mathcal{R_A}(G)
    \end{align}
then the unsupervised error is small:
\begin{align}
    Err_{unsup}(G)\leq \textnormal{max}\left\{ \frac{b}{b-1},2 \right\}\mathcal{R_A}(G)
\end{align}
    
\end{lemma}
$\textit{Proof}$. Using Lemma (\ref{le:2}), $P$ must satisfy $\left(  \frac{\mathcal{R_A}(G)}{b-1}, \mathcal{R_A}(G)\right)$-constant-expansion. By recalling $\textnormal{min}_i P (\left\{ x:G(x=i) \right\}) > \textnormal{max}\left\{ \frac{2}{b-1},2 \right\}\mathcal{R_A}(G)$, we can now apply assumption (\ref{as:3}) to conclude there exist permutation $\pi:[K]\rightarrow[K]$ for any $b>1$ s.t. 

\begin{align}
    P(\left\{ x:\pi(G(x))\neq G^*(x) \right\}) \le \textnormal{max}\left\{ \frac{b}{b-1},2 \right\}\mathcal{R_A}(G)
\end{align}

This shows that our self-learning strategy leveraging consistency regularization in the form of entropy minimization and transductive label propagation loss has an upper bound on the target error.

\subsection{Proofs for Theorem 2}
To prove theorem (\ref{th:2}), we follow the proof by \cite{ge2023provableadvantageunsupervisedpretraining} by reformulating
the contrastive learning task into a maximum likelihood estimation task. We reformulate the contrastive learning task in equation (\ref{eq:cl}) as Gibbs distribution for the paired data as
\begin{align}
    p_{\theta}(f_{\theta}(x_{ij}),s_{ij})= \frac{\exp{(q.k_{+}/\tau)}}{\sum_{j\in\mathcal{N}_{q}}\exp{(q.k_j/\tau)}}
\end{align}
where $f_{\theta}(x_{ij})=(z_i,z_j)=(f_{\theta}(x_i),f_{\theta}(x_j))$ and 
\begin{align}
    \hat{f_{\theta}}\leftarrow\underset{f_{\theta} \in \Theta}{\textnormal{argmax}}\sum_{i,j=1}^{n}\textnormal{log } p_{\theta}(f_{\theta}(x_{ij}),s_{ij})
\end{align}
Using the definition of $\hat{f_{\theta}}$, we have
\begin{align}
\begin{split}
    0&\le \frac{1}{2}\left( \sum_{i,j=1}^{n}log{p_{\hat{{f_{\theta}}}}(x_{ij})}- \sum_{i,j=1}^{n}log{p_{f_{\theta}^*}(x_{ij})}\right)
\end{split}\\
\begin{split}\label{eq:f-theta}
    &= \frac{1}{2} \sum_{i,j=1}^{n}log{\frac{p_{\hat{{f_{\theta}}}}(x_{ij})}{p_{f_{\theta}^*}(x_{ij})}}
\end{split}     
\end{align}
Then we use Markov inequality and Boole inequality to construct relationship between $d_{TV}$ and previous formula. Please note that $\mathcal{P}_{\mathcal{X}\times \mathcal{S}}(\Theta):=\left\{ p_{f_{\theta}}(x,s)|f_{\theta}\in \Theta \right\}$, $\mathcal{N}_{[]}(\mathcal{P}_{\mathcal{X}\times \mathcal{S}}(\Theta),\epsilon)$ be the smallest $\epsilon$-bracket $\mathcal{P}_{\mathcal{X}\times \mathcal{S}}(\Theta)$, and $N_{[]}(\mathcal{P}_{\mathcal{X}\times \mathcal{S}}(\Theta),\epsilon) = \left| \mathcal{N}_{[]}(\mathcal{P}_{\mathcal{X}\times \mathcal{S}}(\Theta),\epsilon) \right|$. For any $\overline{p}_{f_{\theta}}\in  \mathcal{N}_{[]}(\mathcal{P}_{\mathcal{X}\times \mathcal{S}}(\Theta),\epsilon)$, we have the following Markov inequality,

\begin{multline}
    {\scriptstyle
    \mathbb{P}\left( \textnormal{exp}\left( \frac{1}{2}\sum_{i,j=1}^{n}\textnormal{log}\frac{\overline{p}_{f_{\theta}}(x_{ij})}{p_{f_{\theta}^*}(x_{ij})} \ge a\right) \right)\le \frac{\mathbb{E}\left[ \textnormal{exp}\left( \frac{1}{2}\sum_{i,j=1}^{n}\textnormal{log}\frac{\overline{p}_{f_{\theta}}(x_{ij})}{p_{f_{\theta}^*}(x_{ij})} \right) \right]}{a}
    }
\end{multline}
\begin{multline}\label{eq:markov}
    {\scriptstyle
    \mathbb{P}\left( \textnormal{exp}\left( \frac{1}{2}\sum_{i,j=1}^{n}\textnormal{log}\frac{\overline{p}_{f_{\theta}}(x_{ij})}{p_{f_{\theta}^*}(x_{ij})} \ge \frac{\delta\mathbb{E}\left[ \textnormal{exp}\left( \frac{1}{2}\sum_{i,j=1}^{n}\textnormal{log}\frac{\overline{p}_{f_{\theta}}(x_{ij})}{p_{f_{\theta}^*}(x_{ij})} \right) \right]}{C} \right) \right)\le C/\delta
    }
\end{multline}

Define the event $A_{\overline{p}_{f_{\theta}}}$ as
\begin{multline}{\scriptstyle
    A_{\overline{p}_{f_{\theta}}} = \left \{ x:\textnormal{exp} \left ( \frac{1}{2}\sum_{i,j=1}^{n}\textnormal{log}\frac{\overline{p}_{f_{\theta}}(x_{ij})}{p_{f_{\theta}^*}(x_{ij})} \right) \ge \frac{\delta\mathbb{E}\left[ \textnormal{exp}\left( \frac{1}{2}\sum_{i,j=1}^{n}\textnormal{log}\frac{\overline{p}_{f_{\theta}}(x_{ij})}{p_{f_{\theta}^*}(x_{ij})} \right) \right]}{C} \right \}}
\end{multline}
by iterating over all $\overline{p}_{f_{\theta}}\in \mathcal{N}_{[]}(\mathcal{P}_{\mathcal{X}_{\mathcal{T}}\times \mathcal{S}}(\Theta),\epsilon)$, we have
\begin{multline}{\scriptstyle
    \mathbb{P}(\cup_{\overline{p}_{f_{\theta}}\in \mathcal{N}_{[]}(\mathcal{P}_{\mathcal{X}_{\mathcal{T}}\times \mathcal{S}}(\Theta),\epsilon)} A_{\overline{p}_{f_{\theta}}} )\le \sum_{\overline{p}_{f_{\theta}}\in \mathcal{N}_{[]}(\mathcal{P}_{\mathcal{X}_{\mathcal{T}}\times \mathcal{S}}(\Theta),\epsilon)}\mathbb{P}(A_{\overline{p}_{f_{\theta}}})}
\end{multline} 
\begin{multline}{\scriptstyle
    \le \frac{N_{[]}(\mathcal{P}_{\mathcal{X}_{\mathcal{T}}\times \mathcal{S}}(\Theta),\epsilon) \cdot C}{\delta} }
\end{multline}
Take $C=N_{[]}(\mathcal{P}_{\mathcal{X}\times \mathcal{S}}(\Theta),\epsilon)$, then it holds the probability at least $1-\delta$ for all $\overline{p}_{f_{\theta}}\in \mathcal{N}_{[]}(\mathcal{P}_{\mathcal{X}\times \mathcal{S}}(\Theta),\epsilon)$
\begin{multline}{\scriptscriptstyle
    \textnormal{exp}\left( \frac{1}{2}\sum_{i,j=1}^{n}\textnormal{log}\frac{\overline{p}_{f_{\theta}}(x_{ij})}{p_{f_{\theta}^*}(x_{ij})} \right) \le \mathbb{E}\left[ \textnormal{exp}\left( \frac{1}{2}\sum_{i,j=1}^{n}\textnormal{log}\frac{\overline{p}_{f_{\theta}}(x_{ij})}{p_{f_{\theta}^*}(x_{ij})} \right) \right] \cdot  \frac{\mathcal{N}_{[]}(\mathcal{P}_{\mathcal{X}\times \mathcal{S}}(\Theta),\epsilon)}{\delta}}
\end{multline}
\begin{multline}{\scriptscriptstyle
     \frac{1}{2}\sum_{i,j=1}^{n}\textnormal{log}\frac{\overline{p}_{f_{\theta}}(x_{ij})}{p_{f_{\theta}^*}(x_{ij})} \le \textnormal{log }\mathbb{E}\left[ \textnormal{exp}\left( \frac{1}{2}\sum_{i,j=1}^{n}\textnormal{log}\frac{\overline{p}_{f_{\theta}}(x_{ij})}{p_{f_{\theta}^*}(x_{ij})} \right) \right] + \textnormal{log } \frac{\mathcal{N}_{[]}(\mathcal{P}_{\mathcal{X}\times \mathcal{S}}(\Theta),\epsilon)}{\delta}}
\end{multline}
then by referring to the definition of $\hat{f}_{\theta}$ as in equation (\ref{eq:f-theta}) we have
\begin{multline}{\scriptstyle
    0 \le \textnormal{log }\mathbb{E}\left[ \textnormal{exp}\left( \frac{1}{2}\sum_{i,j=1}^{n}\textnormal{log}\frac{\overline{p}_{f_{\theta}}(x_{ij})}{p_{f_{\theta}^*}(x_{ij})} \right) \right] + \textnormal{log } \frac{\mathcal{N}_{[]}(\mathcal{P}_{\mathcal{X}\times \mathcal{S}}(\Theta),\epsilon)}{\delta}}
\end{multline}
\begin{multline}{\scriptstyle
    =\sum_{i,j=1}^{n}\textnormal{log }\mathbb{E}\left[ \sqrt{ \frac{\overline{p}_{f_{\theta}}(x_{ij})}{p_{f_{\theta}^*}(x_{ij})}} \right] + \textnormal{log } \frac{\mathcal{N}_{[]}(\mathcal{P}_{\mathcal{X}\times \mathcal{S}}(\Theta),\epsilon)}{\delta}}
\end{multline}
\begin{multline}{\scriptstyle
    = n^2 \textnormal{ log }\int \sqrt{ \overline{p}_{f_{\theta}}(x)\cdot p_{f_{\theta}^*}(x)}dx  + \textnormal{log } \frac{\mathcal{N}_{[]}(\mathcal{P}_{\mathcal{X}\times \mathcal{S}}(\Theta),\epsilon)}{\delta}}
\end{multline}
\begin{multline}{\scriptstyle
    \le n^2 \left( \int \sqrt{ \overline{p}_{f_{\theta}}(x)\cdot p_{f_{\theta}^*}(x)}dx-1 \right)  + \textnormal{log } \frac{\mathcal{N}_{[]}(\mathcal{P}_{\mathcal{X}\times \mathcal{S}}(\Theta),\epsilon)}{\delta}}
\end{multline}
By rearranging the term, we have
\begin{multline}{\scriptstyle
    1-\int \sqrt{ \overline{p}_{f_{\theta}}(x)\cdot p_{f_{\theta}^*}(x)}dx \le \frac{1}{n^2} + \textnormal{log } \frac{\mathcal{N}_{[]}(\mathcal{P}_{\mathcal{X}\times \mathcal{S}}(\Theta),\epsilon)}{\delta}}
\end{multline}
\begin{multline}\label{eq:A.27} {\scriptstyle
    \int \left(\sqrt{ \overline{p}_{f_{\theta}}(x)} - \sqrt{ p_{f_{\theta}^*}(x)} \right)^2 dx\le \frac{2}{n^2} + \textnormal{log } \frac{\mathcal{N}_{[]}(\mathcal{P}_{\mathcal{X}\times \mathcal{S}}(\Theta),\epsilon)}{\delta}}
\end{multline}
By the definition of $\epsilon$-bracket we obtain
\begin{multline}{\scriptstyle
    \int \left(\sqrt{ \overline{p}_{f_{\theta}}(x)} + \sqrt{ p_{f_{\theta}^*}(x)} \right)^2 dx \le 2 + 2 \int \sqrt{\overline{p}_{f_{\theta}}(x)\cdot p_{f_{\theta}^*}(x)}dx }
\end{multline}
\begin{multline}{\scriptstyle
    \le 2+\int \overline{p}_{f_{\theta}}(x)+ p_{f_{\theta}^*}(x)dx }
\end{multline}
\begin{multline}\label{eq:A.30} {\scriptstyle
    \le 2+2(\epsilon+1) = 2\epsilon+4 }
\end{multline}
Using $d_{TV}$ by Cauchy-Schwarz inequality, it then holds that
\begin{multline}{\scriptstyle
    d_{TV}(\mathbb{P}_{\hat{f_{\theta}}}(x),\mathbb{P}_{f_{\theta}^*}(x)) = \frac{1}{2} \int | p_{\hat{f_{\theta}}}(x)-p_{f_{\theta}^*}(x) |dx }
\end{multline}
\begin{multline}{\scriptstyle
    \le \frac{1}{2} \int | \overline{p}_{\hat{f_{\theta}}}(x)-p_{f_{\theta}^*}(x) |dx + \frac{1}{2} \int | p_{\hat{f_{\theta}}}(x)-\overline{p}_{\hat{f}_{\theta}}(x) |dx }
\end{multline}
\begin{multline}\label{eq:A.33} {\scriptstyle
    \le \frac{1}{2} \left( \int \left( \sqrt{\overline{p}_{\hat{f_{\theta}}}(x)}
    - \sqrt{p_{f_{\theta}^*}(x)}\right)^2 dx \cdot \left( \sqrt{\overline{p}_{\hat{f_{\theta}}}(x)}
    + \sqrt{p_{f_{\theta}^*}(x)}\right)^2 dx 
    \right)^{\frac{1}{2}}+\frac{\epsilon}{2} }
\end{multline}
By substituting equation (\ref{eq:A.27}) and (\ref{eq:A.30}) into equation (\ref{eq:A.33}), we have

\begin{multline}{\scriptstyle
    d_{TV}(\mathbb{P}_{\hat{f_{\theta}}}(x),\mathbb{P}_{f_{\theta}^*}(x)) \le \frac{1}{2} \sqrt{\frac{2}{n^2} + \textnormal{log } \frac{\mathcal{N}_{[]}(\mathcal{P}_{\mathcal{X}\times \mathcal{S}}(\Theta),\epsilon)}{\delta}\cdot (2\epsilon+4)} + \frac{\epsilon}{2}}
\end{multline}

set $\epsilon=\frac{1}{n^2}$, then we have
\begin{multline}{\scriptstyle
    d_{TV}(\mathbb{P}_{\hat{f_{\theta}}}(x),\mathbb{P}_{f_{\theta}^*}(x)) \le 3 \sqrt{\frac{1}{n^2} + \textnormal{log } \frac{\mathcal{N}_{[]}(\mathcal{P}_{\mathcal{X}\times \mathcal{S}}(\Theta),\frac{1}{n^2})}{\delta}}}
\end{multline}
This shows that with contrastive learning the total variation distance between the best feature representation $f_{\theta}^*$
and the learned feature representation $\hat{f_{\theta}}$ can be bounded by the bracket number
of the possible distribution space $P_{\mathcal{X}\times \mathcal{S}}(\Theta)$.

\section{Complexity Analysis}
This section discusses the complexity analysis of the proposed method. Suppose that ${N}$ is the total number of samples of the target dataset consists of B number of batches that satisfies $\sum_{b=1}^{B}N_{b}=N$, $E$ is the number of training epoch, $T$ is the size of temporal queue where $T < N$, and $M$ is the size of memory bank where $M<N$. Following pesudo-code in Algorithm 1, TFDA method consists of several processes i.e. pseudo label refinement, sample selection and computing confidence threshold, uncertainty threshold, and losses that are conducted with complexity $O(1)$, negative pair exclusion, temporal queue and memory bank update. Let $C$ denote the complexity of a process. Following Algorithm 1, the complexity of the proposed method can be written as follows:
\begin{equation}
\begin{split}
    C(TFDA) = C(PseudoLabelRefinement)+C(\tau_{c})+\\C(\tau_{u})+C(SampleSelection)+C(\mathcal{L}_{CE})+\\C(\mathcal{L}_{lp})+C(NegativePairExclusion)+\\C(\mathcal{L}_{CL})+C(\mathcal{L}_{cons})+C(\mathcal{L}_{ul})+\\C(UpdateTemporalQueue)+\\C(UpdateMemoryBank)
\end{split}
\end{equation}
\begin{equation}
\begin{split}
    C(TFDA) = C(PseudoLabelRefinement) + C(\tau_{c}) +\\ C(\tau_{u}) +  C(SampleSelection) +  C(\mathcal{L}_{CE}) +\\ C(\mathcal{L}_{lp}) +  C(NegativePairExclusion) +\\ C(\mathcal{L}_{CL}) + C(\mathcal{L}_{cons}) +  C(\mathcal{L}_{ul}) +\\  C(UpdateTemporalQueue) +\\    C(UpdateMemoryBank)
\end{split}
\end{equation}

\begin{equation}
    \begin{split}
        C(TFDA) = E.\sum_{b=1}^{B} N_{b}(O(M)+ O(1)+ O(1)+O(1)+\\ O(1)+ 
        O(1) +O(T)+O(1)+ \\ O(1) + O(1)+ O(T)+O(M)) \\
        C(TFDA)  = O(E.M.\sum_{b=1}^{B} N_{b})
        +O(E.T.\sum_{b=1}^{B} N_{b})+\\  O(E.T.\sum_{b=1}^{B} N_{b})+\\O(E.M.\sum_{b=1}^{B} N_{b}))
    \end{split}    
\end{equation}
since $\sum_{b=1}^{B} N_{b}=N$ and T is a small number ($T < 10$), then the complexity of TFDA can be written as:
\begin{equation}
    \begin{split}
        C(TFDA) = O(E.M.N)
        +O(E.T.N)+\\O(E.T.N)+O(E.M.N)\\
        C(TFDA)= O(E.M.N)
    \end{split}    
\end{equation}

\begin{algorithm}
    \caption{TFDA}
    \label{algorithm}
    \textbf{Input:}  Target dataset \textit{\(D^{T}=\left \{x_{i}^{T} \right \}_{i=1}^{N_{T}}\)  }, number of samples \( N_{T} \), number of epochs \textit{\( E \)}, number of batch \textit{\( B \)}, size of temporal queue \textit{\( T \)}, size of memory bank \textit{M}\\
    \textbf{Output:} Configuration of \textbf{TFDA} 
    \begin{algorithmic}[1]
		\State Initialize candidates of neighbours $ \{ z'_{j}, p'_{j} \}_{j=1}^{M} $ in memory bank
        \For {\textit{e} = 1 : \textit{E}}
            \For{\textit{b} = 1 : \textit{B}}
                \For{\textit{m} = 1 : \textit{M}}
                    \State Pseudo\_label\_refinement($x_{i}^{T}$, memory bank)
                \EndFor
                \State Compute confidence and uncertainty thresholds ($\tau_{c}, \tau_{u}$) 
                
                \State Sample selection based on $\tau_{c}$ and $\tau_{u}$ 
                \State Compute Cross Entropy loss $\mathcal{L}_{CE}$
                \State Compute Label Propagation loss $\mathcal{L}_{lp} \leftarrow eq.(5)$
                \For{\textit{t} = 1 : \textit{T}}
                    \State Negative\_pair\_exclusion($x_{i}^{T}$, temporal queue)
                \EndFor
                \State Compute Contrastive loss $\mathcal{L}_{CL} \leftarrow eq.(10)$
                \State Compute Self-Distillation loss $\mathcal{L}_{cons} \leftarrow eq.(11)$
                \State Compute Uncertainty Reduction loss $\mathcal{L}_{ul} \leftarrow eq.(12)$
                \For{\textit{t} = 1 : \textit{T}}
                    \State Update temporal queue
                \EndFor
                \For{\textit{m} = 1 : \textit{M}}
                    \State Update memory bank
                \EndFor
            \EndFor		
		\EndFor
    \end{algorithmic} 
\end{algorithm} 

\nomenclature{\(f_{\theta}\)}{Time encoder}
\nomenclature{\(g_{\phi}\)}{Time classifier}
\nomenclature{\(f_{\theta_f}\)}{Frequency encoder}
\nomenclature{\(g_{\phi_f}\)}{Frequency classifier}
\nomenclature{\(\mathcal{X}\)}{Time input space}
\nomenclature{\(\mathcal{X}_F\)}{Frequency input space}
\nomenclature{\(\mathcal{D}^S\)}{Source domain}
\nomenclature{\(\mathcal{D}^T\)}{Target domain}
\nomenclature{\(\mathcal{Y}^S\)}{Source domain label space}\
\nomenclature{\(\mathcal{Y}^T\)}{Target domain label space}
\nomenclature{\(\mathcal{P}^T\)}{Target domain distribution}
\nomenclature{\(\mathcal{P}^S\)}{Source domain distribution}
\nomenclature{\(\mathcal{Z}\)}{Time feature}
\nomenclature{\(\mathcal{Z}_F\)}{Frequency feature}
\nomenclature{\(\mathcal{P}\)}{Time projector}
\nomenclature{\(\mathcal{P}_F\)}{Frequency projector}
\nomenclature{\(\mathcal{A}(x)\)}{Augmentation function of input $x$}
\nomenclature{\(\Theta^{st}\)}{Student model}
\nomenclature{\(\Theta^{te}\)}{Teacher model}
\nomenclature{\(\mathcal{T}_{wa}\)}{Weak augmented data distribution}
\nomenclature{\(\mathcal{T}_{sa}\)}{Strong augmented data distribution}
\nomenclature{\(\mathcal{N}(x)\)}{Neighborhood of $x$}
\nomenclature{\(F\)}{Feature extractor}
\nomenclature{\(G\)}{Classifier}
\nomenclature{\(\mathcal{Q}_i\)}{set of examples with ground-truth label $i$}
\nomenclature{\(N_{[]}(.,.)\)}{bracket number}
\nomenclature{\(\mathcal{R_A}\)}{Consistency regularization loss}

\printnomenclature




\bibliographystyle{model1-num-names}

\bibliography{cas-refs-sfda}





\end{document}